\documentclass[10pt,twocolumn,letterpaper]{article}

\usepackage[pagenumbers]{cvpr} % To force page numbers, e.g. for an arXiv version

\makeatletter
\@namedef{ver@everyshi.sty}{}
\makeatother 

\usepackage{booktabs}
\usepackage{mathptmx}
\usepackage{graphicx}
\usepackage{amsmath}
\usepackage{amsthm}
\usepackage{amssymb}
\usepackage{wrapfig}

\usepackage[pagebackref=true,breaklinks=true,colorlinks,bookmarks=false]{hyperref}

\usepackage{xcolor}
\definecolor{fb_blue}{RGB}{66 103 178}
\definecolor{fb_grey}{RGB}{137 143 156}
\definecolor{ins_yellow}{RGB}{255,220,128}
\definecolor{ins_orange}{RGB}{247,119,55}

\xdefinecolor{navyblue}{RGB}{19 41 75}
\xdefinecolor{carolinablue}{RGB}{123 175 212}
\xdefinecolor{cb}{RGB}{123 175 212}
\xdefinecolor{flatgrey}{RGB}{162 170 173}
\xdefinecolor{matcha}{RGB}{183 224 176}
  
\usepackage[capbesideposition=right]{floatrow}

\usepackage[linesnumbered,ruled,vlined]{algorithm2e}

\SetCommentSty{mycommfont}
\SetKwInput{Input}{Input}                
\SetKwInput{Output}{Output}         
\SetKwInput{Dataset}{Dataset}     
\SetKwInput{Settings}{Settings}      
\SetKwInput{Initialization}{Initialization}      
\SetAlgorithmName{Alg.}{Alg.}{List of Algorithms} 
\makeatletter
\newcommand{\algorithmfootnote}[2][\footnotesize]{%
  \let\old@algocf@finish\@algocf@finish% Store algorithm finish macro
  \def\@algocf@finish{\old@algocf@finish% Update finish macro to insert "footnote"
    \leavevmode\rlap{\begin{minipage}{\linewidth}
    #1#2
    \end{minipage}}%
  }%
}
\makeatother

\usepackage{floatrow}
\usepackage{graphicx}
\usepackage{booktabs}
\usepackage{multirow}
\usepackage{hyperref} 
  
\usepackage{url}

\definecolor{olive}{rgb}{0.42, 0.56, 0.14}

\newcommand{\ccc}[1]{{\textcolor{black}{#1}}}

\newcommand{\Enc}{\small{ \mbox{\fontfamily{qcs}\selectfont{Enc}}}}
\newcommand{\Dec}{\small{ \mbox{\fontfamily{qcs}\selectfont{Dec}}}}
\newcommand{\Dm}{\small{ \mbox{\fontfamily{qcs}\selectfont{Dm}}}}
\newcommand{\Fr}{\small{ \mbox{\fontfamily{qcs}\selectfont{Fr}}}}
\newcommand{\Gs}{\small{ \mbox{\fontfamily{qcs}\selectfont{Gs}}}}
\newcommand{\Kp}{\small{ \mbox{\fontfamily{qcs}\selectfont{Kp}}}}
\newcommand{\yaw}{\small{ \mbox{\fontfamily{qcs}\selectfont{yaw}}}}
\newcommand{\yaws}{\scriptsize{ \mbox{\fontfamily{qcs}\selectfont{yaw}}}}
\newcommand{\Supunsup}{$\small{ \mbox{\fontfamily{qcs}\selectfont{Sup+Unsup}}}$}
\newcommand{\Sup}{$\small{ \mbox{\fontfamily{qcs}\selectfont{Sup}}}$}
\newcommand{\Unsup}{$\small{ \mbox{\fontfamily{qcs}\selectfont{Unsup}}}$}

\newcommand{\nsrc}{N}
\newcommand{\nexpr}{M}
\newcommand{\nkp}{K}

\definecolor{babypink}{rgb}{0.96, 0.76, 0.76}
\definecolor{mygreen}{rgb}{0.0, 0.5, 0.0}

\def\genbox#1#2#3#4#5#6{% #1=0/1, #2=color, #3=shape, #4=raise, #5=width, #6=width/2
    \leavevmode\raise#4bp\hbox to#5bp{\vrule height#5bp depth0bp width0bp
    \pdfliteral{q .5 w \csname #2COLOR\endcsname\space RG
                       \csname #3PDF\endcsname{#5}{#6} S Q
             \ifx1#1 q \csname #2COLOR\endcsname\space rg 
                       \csname #3PDF\endcsname{#5}{#6} f Q\fi}\hss}}

\def\trianbox   #1#2{\genbox{#1}{#2}  {trian}    {0}   {5}    {2.5}}
\def\uptrianbox #1#2{\genbox{#1}{#2}  {uptrian}  {0}   {5}    {2.5}}

\newcommand{\spacefill}[1]{ \scriptsize \textcolor{white}{\trianbox1{white} {#1}$\%$}}
\newcommand{\improv}[1]{ \scriptsize \textcolor{mygreen}{\trianbox1{cgreen} {#1}$\%$}}
\newcommand{\down}[1]{ \scriptsize \textcolor{babypink}{\uptrianbox1{cpink} {#1}$\%$}}

\usepackage{caption}
\usepackage{subcaption}

\begin{document}

%%%%%%%%% TITLE
\title{Efficient conditioned face animation using frontally-viewed embedding}
% for low bandwidth video chat

\author{Maxime Oquab$^{\star}$, Daniel Haziza$^{\star}$, Ludovic Schwartz$^{\dagger}${\thanks{Work achieved during internship at Meta.}}, Tao Xu$^{\star}$,\\ Katayoun Zand$^{\star}$, Rui Wang$^{\star}$, Peirong Liu$^{\diamond *}$, Camille Couprie$^{\star}$\\
$^\star$ Meta, $^\dagger$ Universidad Pompeu Fabra, $^\diamond$ UNC-Chapel Hill \\
{\tt\scriptsize{\{qas, dhaziza, xutao, coupriec\}@fb.com}}}
\maketitle

%%%%%%%%%%%%%%%%%%%%%%%%%
%\input{sec0-abstract.tex}
\begin{abstract}
As the quality of few shot facial animation from landmarks increases, new applications become possible, such as ultra low bandwidth video chat compression with a high degree of realism. However, there are some important challenges to tackle in order to improve the experience in real world conditions. In particular, the current approaches fail to represent profile views without distortions, while running in a low compute regime. We focus on this key problem by introducing a multi-frames embedding dubbed Frontalizer to improve profile views rendering. In addition to this core improvement, we explore the learning of a latent code conditioning generations along with landmarks to better convey facial expressions.   
Our dense models achieves $22\%$ of improvement in perceptual quality and $73\%$ reduction of landmark error over the first order model baseline on a subset of DFDC videos containing head movements. Declined with mobile architectures, our models outperform the previous state-of-the-art (improving perceptual quality by more than $16\%$ and reducing landmark error by more than $47\%$ on two datasets) while running on real time on iPhone 8 with very low bandwidth requirements.    
\end{abstract}

\section{Introduction}

Despite the progress of recent video codecs, they fail to provide meaningful videos in ultra low bandwidth situations. To bridge this gap, recent generative approaches \cite{wang2020oneshot,konuko2020ultralow,oquab2020low} suggested to compress and send facial landmarks with a reference image, and animate it using variants of the First Order Model (FOM) work\cite{Siarohin_2019_NeurIPS}.  

To be used for video chat, a generative low-bandwidth approach needs to satisfy several constraints. Some are related to inter-person communication, {\it{e.g.}} identity preservation without uncanny valley effects, and proper fidelity of facial motion. Some are related to the technical context, {\it{i.e.}} the algorithm must work in real-time, preferably on mobile devices, and use only little bandwidth.
\begin{figure}[ht]
    \centering
    \includegraphics[width=0.98\textwidth]{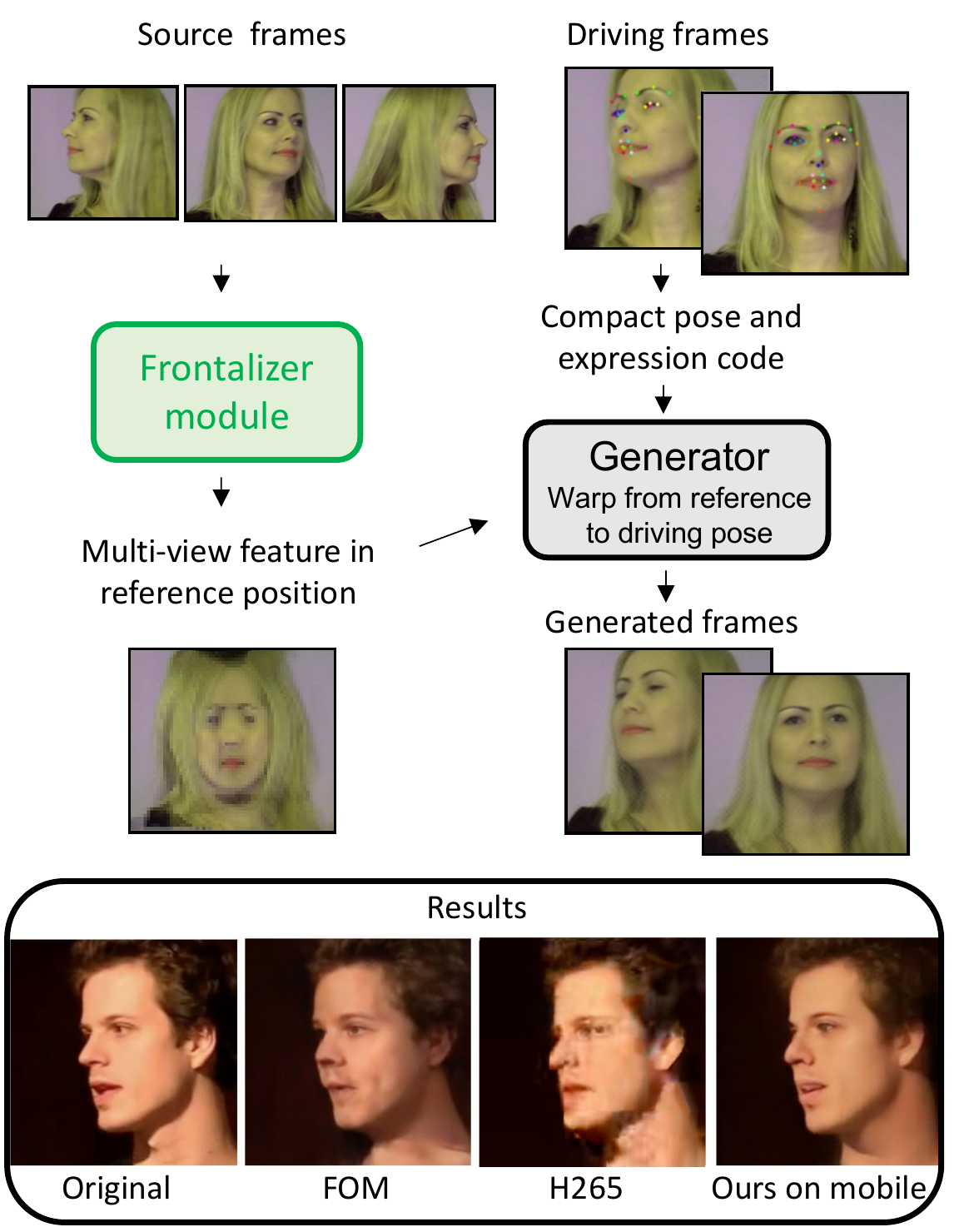} 
    \caption{We introduce the Frontalizer module, that builds a unified embedding from a set of source frames, allowing our generator based on the First Order Model \cite{Siarohin_2019_NeurIPS} (FOM) to use appearance information from multiple frames. While in the algorithm, this happens in \textbf{feature space}, for this figure we apply the same transforms to the source images for illustrating the multi-view feature. Our resulting face animation approach is running on real time on iPhone8 at less than 7 kbps (kbits per second) using three source frames, improving over FOM and the HEVC codec. The transmission of two additional frames is amortized in a few seconds.}
    \label{fig:teaser}
\end{figure}

\begin{figure}[t]
    \centering
    \includegraphics[width=0.98\textwidth, trim=0 10cm 0 0]{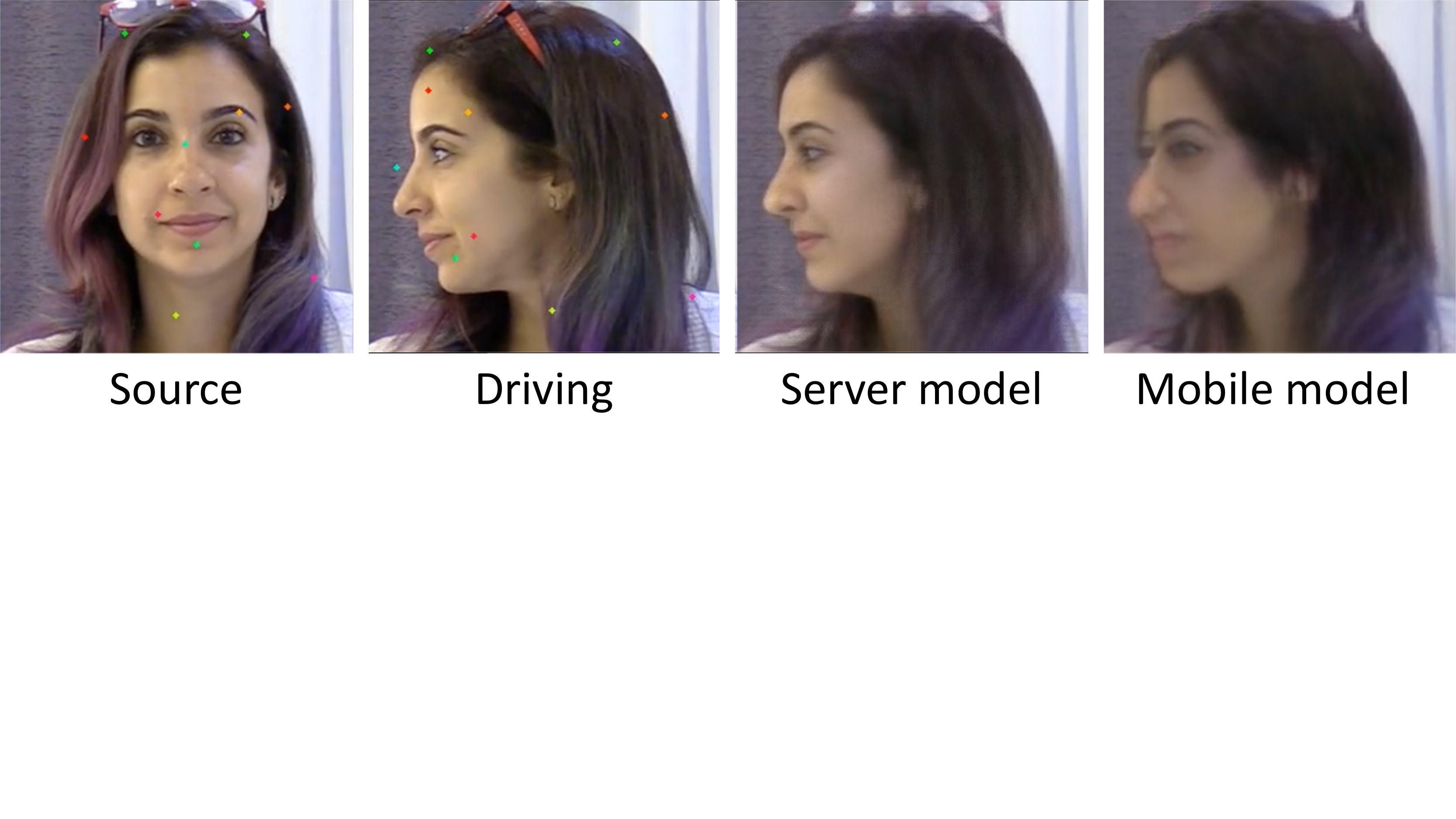}
    \caption{Reconstruction of a side view by a \textbf{baseline FOM} \cite{Siarohin_2019_NeurIPS} \textbf{model} using one frontal source frame. From left to right: source image with keypoints; driving image with keypoints; reconstruction with a dense model; reconstruction with a mobile model. The side-view appearance information is absent, and the results present noticeable distortions. The effect is stronger on mobile where the capacity is low.}
    \label{fig:side-view-problem}
\end{figure}

Leading face animation approaches are based on the FOM work by deforming a source image to match the position of a driving image. However, this approach suffers from one fundamental issue: the hidden areas, that were occluded in the source image, are implicitly inpainted by an image generator model at inference time. This can lead to small perceptible differences in the result and an uncanny valley effect, as it can be seen in the FOM result of Figure~\ref{fig:side-view-problem}. When approaching the problem with mobile models, the reconstruction quality for these areas decreases even more as the capacity of the model is reduced and the result becomes unacceptable for a video call.
In order to tackle this problem in a small model regime, it is critical to use multiple views of a face. Combining these views efficiently is a non-trivial problem. One possible approach would consist of dynamically switching between several source frames and picking the one closest to the driving position. However, for the result to appear natural and temporally consistent, it is important to avoid abrupt changes in the appearance as can happen when switching from one source frame to another. Therefore, our first contribution is the introduction of a Frontalizer module, that combines a set of source frames into a single reference embedding map, containing the information from all views of the source set in feature space, see Figure \ref{fig:teaser}.

% the problem of expressions
Then, in order to refine the result, we introduce a lightweight latent code that is transmitted along with the driving position. This code is given to the generator network and optimized to improve the reconstruction. We show that this can refine difficult areas such as the inside of the mouth.

% the problem of mobile inference
Finally, as the goal of this work is to enable a generative-based video-chat, we propose a mobile architecture and implementation that runs at 30 FPS on a 2017 iPhone 8, leveraging GPU hardware acceleration, while constantly outperforming the state-of-the-art on two datasets.

\section{Related work}

\paragraph*{Facial animation.}

The First Order Model of Siarohin {\it et al.} \cite{Siarohin_2019_NeurIPS} provides a strong baseline for face animation models. Unlike prior work \cite{zakharov2019few, Wiles_2018_ECCV}, it only requires one source frame to accurately reconstruct the frontal view of talking heads. However, solely using one face image makes the reconstruction of profile views extremely challenging, even more so in a low computation regime. 

Different attempts appeared to cope with this problem. In \cite{zakharov2019few}, a finetuning step using a minimum of height frames compensates the identity loss in the approach, that would be tricky to implement on device.
In the Bilayer approach \cite{zakharov2020fast}, the source image is first transformed to extract a high frequency feature embedding, then used to compute the feature deformation field. A drawback of high/low frequency differentiation is the presence of hair artefacts. In \cite{wang2020oneshot}, it is the use of 3D landmarks that aims to provide accurate profile views. However the rotation angle remains limited and the very high dimension of necessary feature space for this approach prevents immediate application to real time on device inference.
Closer to our solution, the work of Wiles {\it et al.} \cite{Wiles_2018_ECCV} embeds multiple frames using a single encoder, but then deforms images directly and not their features, leading to limited performance. 
Audio based approaches such as \cite{Zhou2021poseControlable} animate faces from audio features, a source image and driving poses.   \cite{Zhou2021poseControlable} demonstrates convincing profile views rendering, however the code necessary to condition the generator would be too large (several thousands of floats) to make this approach work in a low bandwidth situation.    
Finally, the very recent work of \cite{liu2021selfappearanceaided} leverages multiple frames by fusing warped features from $N$ multiple views in the driving domain. At inference time, it requires $N$ extra calls to a warping network, making this approach unfit to low compute applications.   

\paragraph*{Expression conditioned models.}

Only conditioning generation on facial landmarks is not enough information to model some facial attributes that can be either occluded  (such as the tongue and teeth) or ignored (e.g. wrinkles) by classical perceptual losses.  
In the literature, we only found approaches conditioning generations on facial expressions \cite{Pumarola2018Ganimation} or facial landmarks (see the previous section), but not both jointly.

\paragraph*{Applications to low bandwidth video chat.}

Last year simultaneously appeared three generative approaches with applications to low bandwidth video chat: \cite{wang2020oneshot,konuko2020ultralow,oquab2020low}. The transfer of Wang {\it et al.}'s approach discussed above \cite{wang2020oneshot} to low compute regime has not been studied so far, but the dimensionality of the latent space employed in their work is \ccc{height times} higher than the one of FOM\ccc{, and 32 times higher than the one of \cite{oquab2020low}. Additionally, the use of 3D convolutions adds another level of complexity that would cause issues in low compute scenarios.} It is therefore unlikely to be a system applicable on device.
A server based solution is evoked in \cite{blognvidia}, where the reconstructed video of a sender with poor network connection could be sent to a receiver to only enable one way communication: The low bandwidth context would prevent the sender to receive their interlocutor's video.
The work of Konuko {\it et al.} \cite{konuko2020ultralow} is fully orthogonal to ours, as it introduces an algorithm to select intra frames to best compress videos using the original FOM algorithm. 
Finally, \cite{oquab2020low} presents a variant of FOM that runs on mobile, where the decoder contains SPADE layers to refine some areas of the face. Like FOM, it does not handle profile views generation well. Moreover, the addition of keypoints required by SPADE augments the  bandwidth.

\section{Model}

\begin{figure*}[ht]
    \centering
    \includegraphics[width=0.95\textwidth]{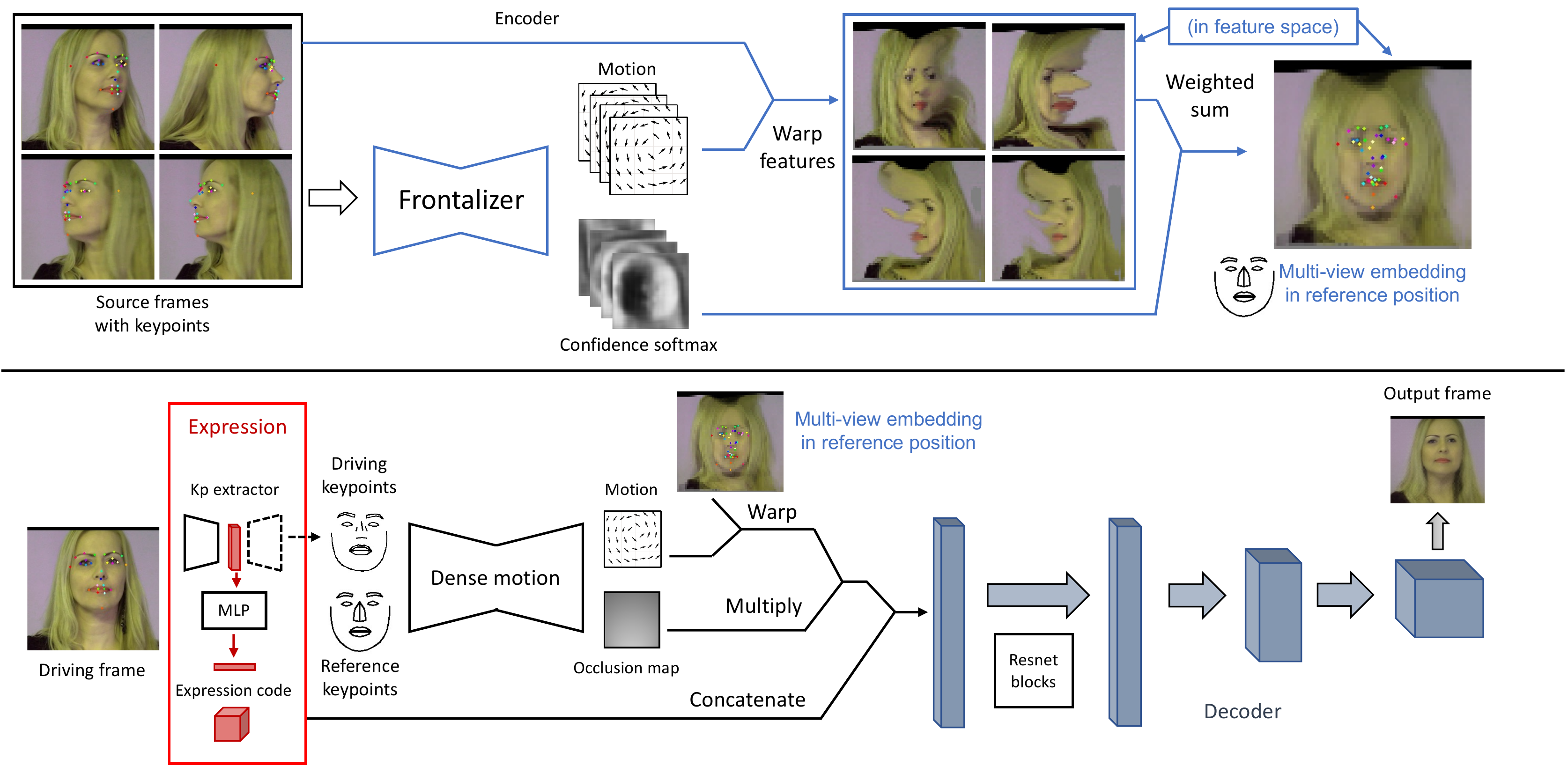} 
    \caption{Overview of our approach. Top part: the Frontalizer first builds a face appearance embedding from a few source images, in a reference position. For the purpose of this figure, we illustrate the features by applying the transforms on the RGB frames to gain visual intuition. In the algorithm, these transforms are applied on the encoder feature maps. Bottom part: We use the multi-view embedding in the reference position to decode the result in a new driving position, hence reconstructing the driving frame only using the driving keypoints and an expression code. Finally, our decoder reconstructs the frame.}
    \label{fig:overview} \label{fig:frontalizer}
\end{figure*}

We describe the model that we propose for low-bandwidth video-chat on mobile phones. In particular we explain which design choices were made to comply with constraints related to the task of low-bandwidth video-chat.

\subsection{Challenges of face animation}

\subsubsection{Identity preservation}
In order to provide a good video-chat experience, it is necessary that the identity of the caller is preserved. While it is difficult to computationally measure identity preservation in image generation, we observe that the First Order Model, based on warping feature maps, offers the best quality on that aspect as it preserves many details of the original face; we build on the FOM approach for the rest of this work.

\paragraph{First Order Model.} We provide a brief description of FOM below and refer the reader to the original paper \cite{Siarohin_2019_NeurIPS} for more details.
For a source frame $x_s$ and a driving frame $x_d$, an unsupervised keypoint detector $\Kp$ computes a set of keypoints and their local Jacobian matrices, defining a source position $p_{s}$ (and similarly, a driving position $p_d$):
$$ p_{s} = \Kp(x_s) \;\;\;\; ; \;\;\;\; p_{d} = \Kp(x_d). $$
An encoder network $\Enc$ computes the feature map embedding for the source frame in position $p_s$:
$$ e^{p_s} = \Enc(x_s).$$
Next, a dense motion network $\Dm$ constructs a flow transformation map $t^{p_s \rightarrow p_d}$ and an occlusion map $o^{p_s \rightarrow p_d}$ in order to build the feature map embedding in position $p_d$:
$$ \left(t^{p_s \rightarrow p_d},\;o^{p_s \rightarrow p_d}\right) = \Dm(x_s, p_s, p_d).$$
Then, the flow is applied onto the feature map using the warping ``grid-sample" operator $\Gs$, before applying the occlusion map by element-wise multiplication $\odot$:
$$ \Tilde{e}^{p_d} = \Gs(e^{p_s}, t^{p_s \rightarrow p_d}) \; \odot \; o^{p_s \rightarrow p_d}. $$
Finally, a decoder network $\Dec$ provides the result image:
$$ \Tilde{x_d} = \Dec(\Tilde{e}^{p_d}).$$

\subsubsection{Inpainting missing areas}
One fundamental problem in warping-based approaches is their reliance on inpainting areas that were not visible in the original image. While it seems natural for a video-chat scenario to use a front-facing source frame, this means that the sides of the face will not be visible. In practice this means that when a head rotation happens, the side-view result image will be partially generated by the model, building on what was seen by the model in the training dataset. At inference time, on a new unseen user, the result will not match, leading to an \textit{uncanny valley effect} (see Figure \ref{fig:side-view-problem}). Moreover, as this implicit inpainting relies on the $capacity$ of the model to learn corresponding patterns in the training dataset, this problem is even more pronounced in a small model regime as is the case for mobile inference purposes, as detailed further in Section \ref{mobile_implementation}. We approach this inpainting problem with our Frontalizer module described in Section \ref{sec:frontalizer}.

\subsubsection{Expression preservation}
One other problem we observe is the one of facial expressions preservation during the video-chat: during communication, facial expressions appear as a combination of motion of facial areas (e.g. eyebrows raising or frowning) and short displays of the corresponding facial wrinkles (e.g. horizontal forehead lines or vertical line between eyebrows). However, flow-based methods such as the FOM or its proposed variant in \cite{wang2020oneshot} (see Figure \ref{fig:wrinkles}) tend to have trouble reproducing the latter. We approach this problem with our expression conditioning setup described in Section \ref{sec:expression-conditioning}.

\subsection{Proposed approach}

An overview of the full pipeline for our approach appears in Figure~\ref{fig:overview}. 
Our work introduces two main components: the Frontalizer module contains a first learnable subnetwork, similar to the dense-motion network: its role is to produce a flow map from a source position to a reference position $p^r$. The reference position $p^r$ corresponds to a set of $\nkp$ keypoint positions; we detail this component in Section~\ref{sec:frontalizer}. 

The second component is an expression module, where a small CNN extracts from the driving frame a compact expression code to refine the result. This expression code is used by the decoder network as conditioning, in the form of one-hot maps. We detail this component in Section~\ref{sec:expression-conditioning}. 

\subsubsection{Frontalized embedding warp and fusion}\label{sec:frontalizer}

The frontalizer module replaces the encoder module of the First Order Model. Given $\nsrc$ source frames $x_1,\dots,x_\nsrc$ in positions $p_1,\dots,p_\nsrc$, its goal is to compute and warp their feature maps into a reference position $p_r$ in feature space, then combine them together depending on their individual confidence maps $c_1,\dots,c_\nsrc$. 
We illustrate this component in the top part of Fig.~\ref{fig:frontalizer}, and detail its steps below:

First, we apply an encoder $\Enc$ to obtain the feature map for each image $x_i$ at the original position $p_i$:
    
    $$ e^{p_i} = \Enc(x_i).$$
Then, we compute a flow map $t_i^{p_i \rightarrow p_r}$ and a confidence map $c_i^{p_i \rightarrow p_r}$ using a dense motion network $\Dm$ re-purposed for frontalization, denoted $\Fr$:
    $$ \left(t^{p_i \rightarrow p_r}, c_i^{p_i \rightarrow p_r}\right) = \Fr\left(x_i, p_i, p_r\right). $$
We apply this flow to the feature maps to obtain the warped result in position $r$, using the grid-sample operator $\Gs$:
    $$ \Tilde{e}_i^{p_r} = \Gs\left(e^{p_i}, t^{p_i \rightarrow p_r} \right). $$
Next, we compute weighting coefficient maps $w_i$ for each flow map by applying a softmax operation to the $\nsrc$ confidence values at each coordinate position $(u,v)$:
 $$ w_{i}(u,v) = \underset{i}{\mbox{softmax}}({c_i}^{p_i \rightarrow p_r}(u,v)) ~\forall i\in\{1,\dots,\nsrc\}.$$ 
Finally, we combine the different maps with an element-wise multiplication together to obtain the fused embedding:
    $$ \Tilde{e}^{p_r} = \sum_{i=1}^\nsrc w_i \; \Tilde{e}_i^{p_r}.$$
This fused embedding $\Tilde{e}^{p_r}$ and its reference position $p_r$ are ready to be used by the generator (dense-motion and decoder networks) of a FOM setup. As it contains information from multiple views, the reliance on the inpainting properties of the decoder is now greatly reduced.
We note that the learnable parts of the Frontalizer module consist of a dense motion network, that specializes in producing flow maps towards a reference position $p_r$ (also learned), as well as corresponding confidence maps, optimizing the merging of the different sources.

\ccc{
\paragraph{Losses.} We use the same loss terms as in the FOM work \cite{Siarohin_2019_NeurIPS}, in particular, the multi-scale perceptual loss and adversarial training.}

\subsubsection{Expression conditioning}\label{sec:expression-conditioning}

To convey important facial expressions such as wrinkles and so on, it is essential to introduce a code to transmit in addition to the landmarks. 

We introduce a vector $e \in \mathbb{R}^{\nexpr}$, of size $\nexpr=16$ by default, aiming to store expression information. We feed this code to our model in input to the decoder. To concatenate its values to the input feature map of the decoder, we fill $E_k$ maps with the spatial repetition of each value $e_k$, $k \in \{ 1, \dots, {\nexpr}\}$. 
To ensure that the given code serves the recovery of desired facial expressions, we introduce a facial action unit loss using predictions from a network pre-trained to predict landmarks with its branch $f_L$, and to classify facial action units with branch $f_E$ ~\cite{toisoul2021estimation,cohn2007observer,yu2012perception}. The loss enforcing expressions is defined as follows:
$$
\mathcal{L}_{E}(x_d, \tilde{x}_d) = \gamma_E ( \|f_E(\tilde{x}_d) - f_E(x_d)\|_1 + \|f_L(\tilde{x}_d) - f_L(x_d)\|_1),  
$$
where $\gamma_E \in\mathbb{R}^+$ and $\tilde{x_d}$ is the output generation by our model from a driving image $x_d$. 
To predict the expression code, we take advantage of the landmark extractor $\Kp$ encoder $f_m$, and define a two-layer Perceptron $g$ on top of it:
$$ e= g (f_m(x_d)).$$
As the learning of $e$ is only supervised by the loss on the final image \ccc{(expression $\mathcal{L}_{E}$ in addition to other terms, e.g. perceptual, adversarial)}, we denote this expression conditioning as free. We also experiment two variants: taking $e = f_E(x_d)$, referred as Oracle, and alternatively further supervising the learning of $e$ using oracle prediction. The free strategy remains the best as shown in our ablation study.

\subsection{Setup details}

\paragraph{Architecture.}
Our models contain several sub-networks that are common with the original FOM work\cite{Siarohin_2019_NeurIPS}: an encoder, a keypoint detector, a dense-motion network and a decoder network. For these networks, we refer the reader to the original paper for their architecture definitions. Architecture details of the Frontalizer and expression module, that are specific to this work, are given in the Appendix. 

\paragraph{Landmark sets.}

We experiment with two variants of keypoint sets: supervised landmarks setups (\Sup) use a set of 33 facial landmarks using a facial landmark tracker. Unsupervised setups (\Unsup) use 10 unsupervised keypoints obtained with the trained keypoint detector, while \Supunsup~use both. Frontalizer models trained only with the \Unsup~set proved difficult to train. Our hypothesis is that the initialization of the reference keypoints positions (see next section) is important, and it is unclear how to initialize the unsupervised set.

\paragraph{Training.}
\label{sec:training}

\begin{figure}[ht]
    \centering
     \includegraphics[width=0.48\textwidth]{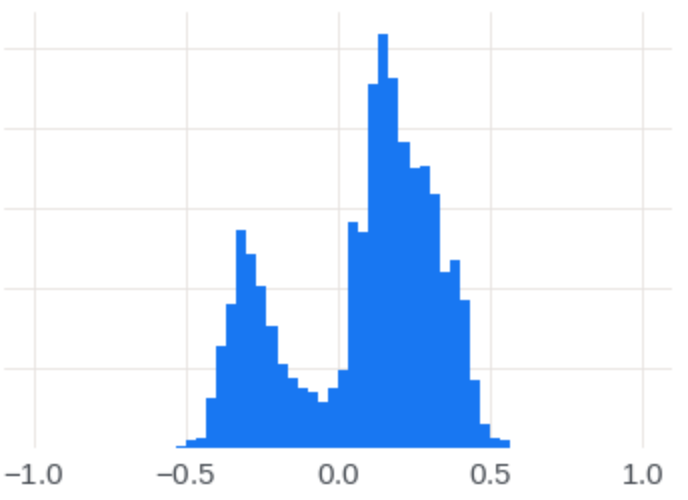}
     \includegraphics[width=0.48\textwidth]{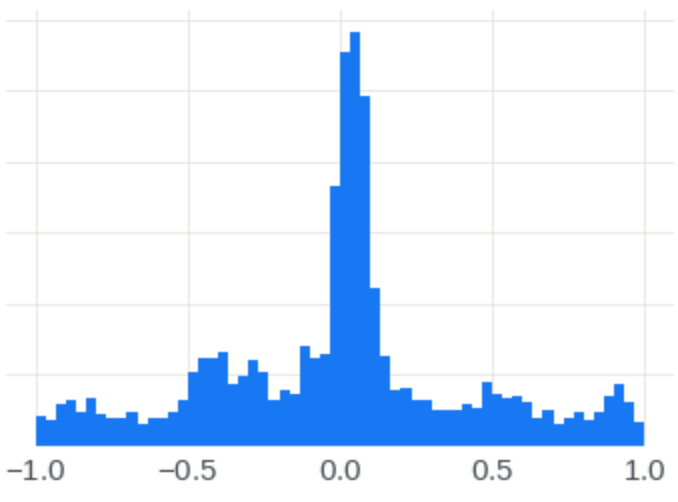}
    \caption{Yaw distributions (radians) for the test videos in VoxCeleb (left) and DFDC (right). Average (across videos) standard deviation of the yaw angle: VoxCeleb: 0.08; DFDC: 0.29. VoxCeleb videos contain only little head rotations compared to DFDC.}
    \label{fig:angle-distribution}
\end{figure}

We use the DFDC dataset \cite{dolhansky2019deepfake} for training, as it provides videos with a wide range of head rotations, which is not the case for the VoxCeleb dataset (see Figure \ref{fig:angle-distribution}).
For training Frontalizer setups, we sample the dataset examples in the following way: given a video of a person talking, we run an off-the-shelf face detector \cite{zhang2017s3fd}, and crop a square box in the video that contains the detected head for over 90 consecutive frames to obtain a \textit{video tube} with a fixed background. We then run a head angle detector on the frames of the tubes, returning their yaw (vertical axis rotation angle). We sample three frames $A,B,C$ where $\yaw_A > \yaw_B > \yaw_C$ and $\left(\yaw_A - \yaw_C\right) > 0.3~{\displaystyle {\text{rad}}}$. We use $A$ and $C$ as source frames, and $B$ as driving frame.

\ccc{While this sampling strategy works well in presence of large head rotations, it neglects identity preservation in videos with less motion. To cope with that, we propose the following improved sampling using four frames $A,B_1,B_2,C$ where 
$$\yaw_A > \yaw_{B_1} > \yaw_C,$$ 
$$\yaw_A - \yaw_C > 0.3~{\displaystyle {\text{rad}}},$$ 
$$\left|\yaw_A - \yaw_{B_2}\right| < 0.1~{\displaystyle {\text{rad}}}.$$
We use $A$ and $C$ as source frames, and $B_1, B_2$ as driving frames.
This sampling allows the model to simultaneously learn from a large rotation regime ($B_1$ far from $A$ and $C$) and a small rotation regime ($B_2$ close to $A$).}

The Frontalizer reference keypoints are initialized, for the part corresponding to supervised facial landmarks, to the average position in the dataset, which is a front-facing layout. The reference unsupervised keypoints are initialized randomly close to the center. These reference keypoint positions are learned end-to-end during training.

\paragraph{Losses parameters.} We use the Adam \cite{kingma2014adam} optimizer ($\beta_1=0.5, \beta_2=0.99$), with all learning rates set to $2.10^{-4}$; we divide the learning rate by 10 after 60k and 90k iterations, and use a batch size of 32 in all our ablation study experiments. We stopped training after 100k iterations, and fine-tune the FOM model with an adversarial loss \cite{goodfellow2014generative} for 50k more iterations as in  \cite{Siarohin_2019_NeurIPS}.

\paragraph{Mobile models.}

\label{mobile_implementation}

In order to run on mobile in real-time, the networks compute requirements and capacity must be greatly reduced. 
As mentioned in the introduction, decreasing the size of the model diminishes the inpainting capabilities of the decoder network. When the capacity is reduced to fit on mobile, this implicit inpainting becomes very difficult in practice - in particular, when there are head rotations - and the Frontalizer approach becomes extremely useful.

To enable mobile inference in real-time, we build all network architectures using efficient MobileNetV2 \cite{sandler2018mobilenetv2} inverted residual building blocks and reduce the size of the central feature from $64\times64$ to $32\times32$. We provide architecture details in the Appendix.

% \input{sec4-experiments.tex}

%----------------------------------------------------------
\section{Experiments}
%----------------------------------------------------------

\begin{figure*}[ht]
    \centering
    \setlength{\tabcolsep}{1pt}
    \begin{tabular}{cccccccc}
    Source $x_1$ & Source $x_2$ & Source $x_3$ & Driving & FOM adv & Front. S & Front. S+U & Front S+U+expr  \\
    \includegraphics[width=0.12\textwidth,height=0.12\textwidth]{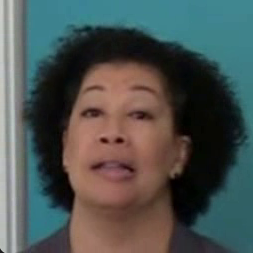}&
    \includegraphics[width=0.12\textwidth,height=0.12\textwidth]{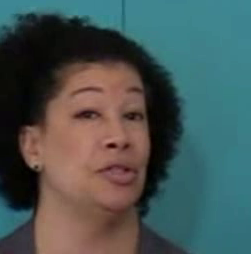}&
    \includegraphics[width=0.12\textwidth,height=0.12\textwidth]{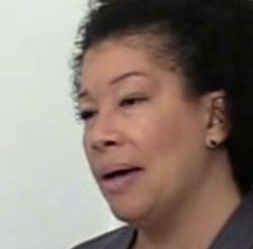}&
    \includegraphics[width=0.12\textwidth,height=0.12\textwidth]{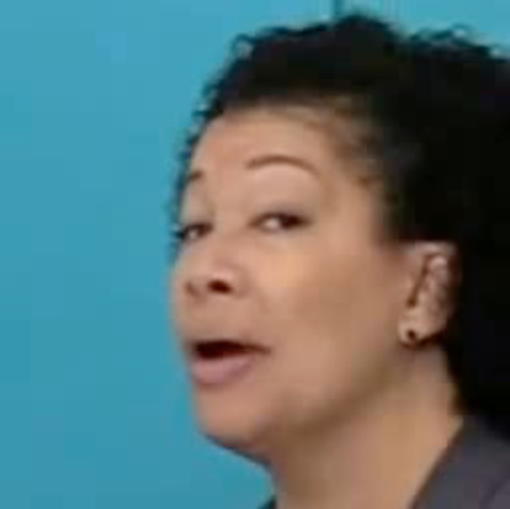}&
    \includegraphics[width=0.12\textwidth,height=0.12\textwidth]{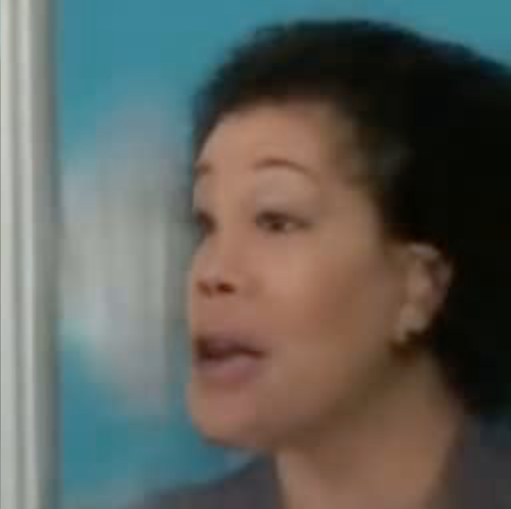}&
    \includegraphics[width=0.12\textwidth,height=0.12\textwidth]{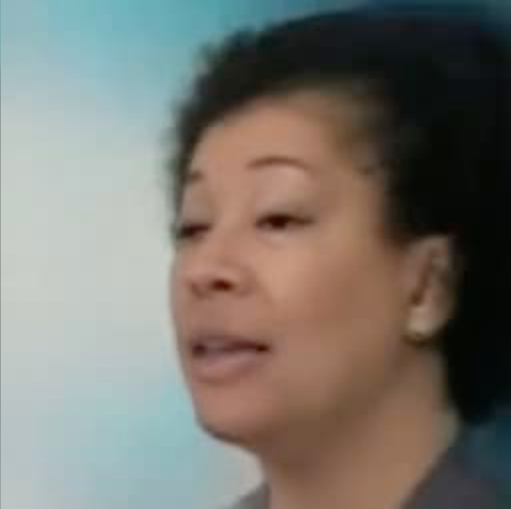}&
    \includegraphics[width=0.12\textwidth,height=0.12\textwidth]{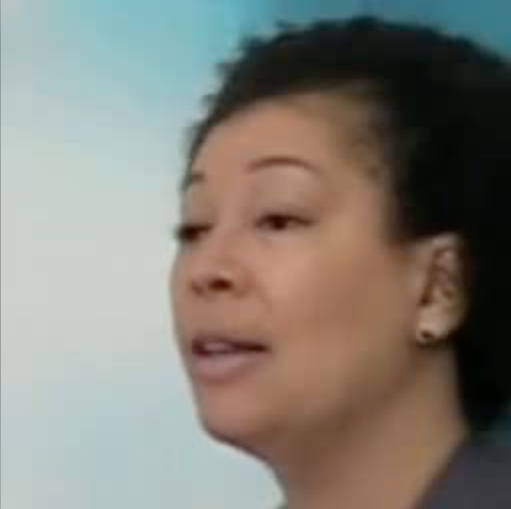}&
    \includegraphics[width=0.12\textwidth,height=0.12\textwidth]{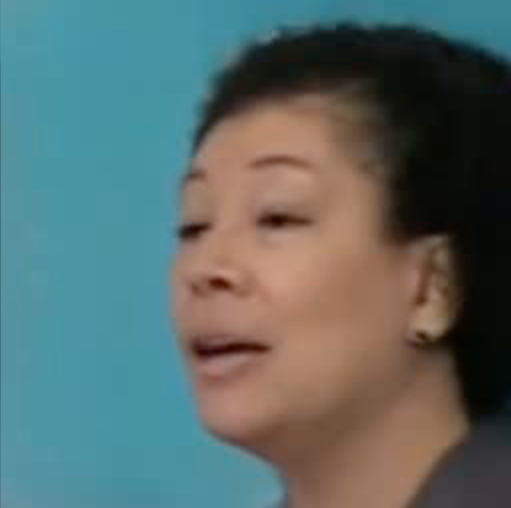}\\
    % row 2
     \includegraphics[width=0.12\textwidth,height=0.12\textwidth]{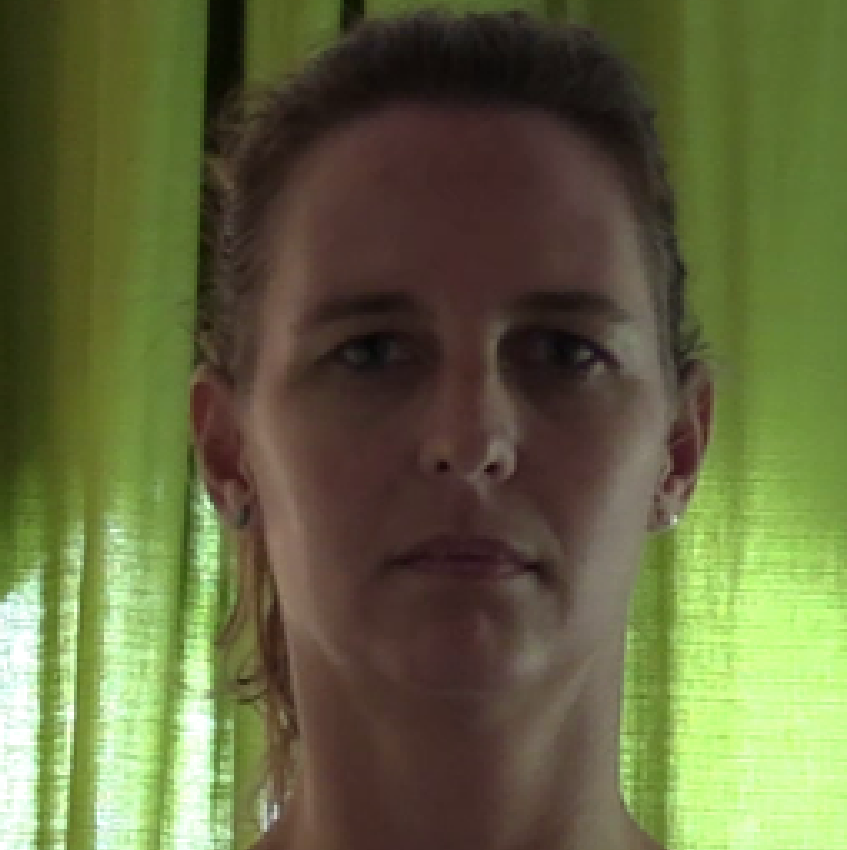}&
    \includegraphics[width=0.12\textwidth,height=0.12\textwidth]{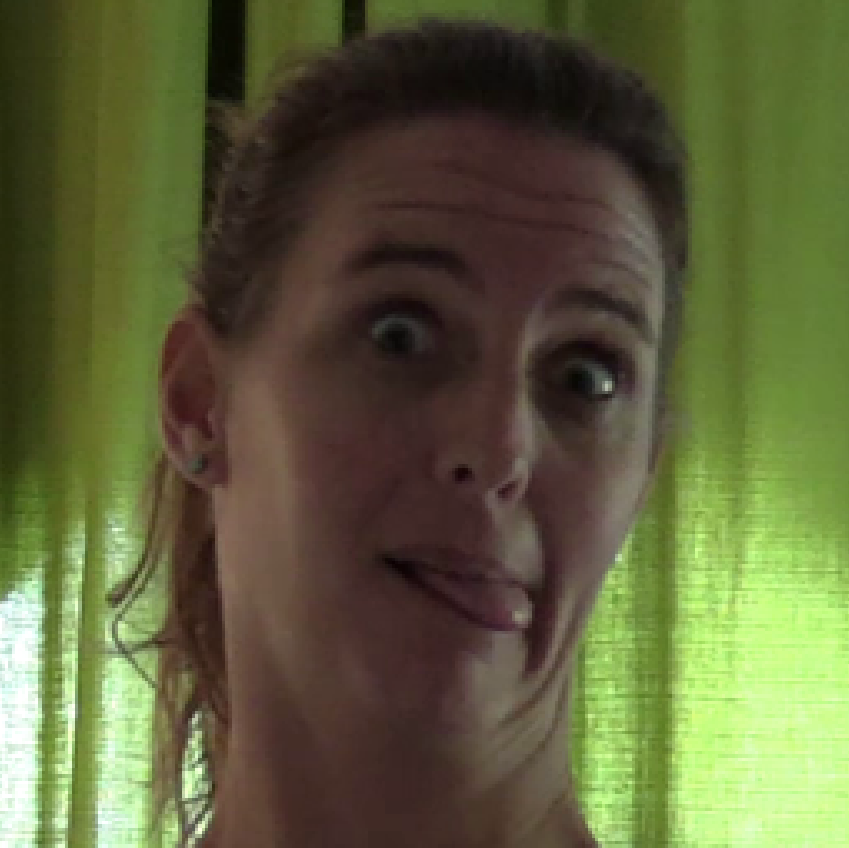}&
    \includegraphics[width=0.12\textwidth,height=0.12\textwidth]{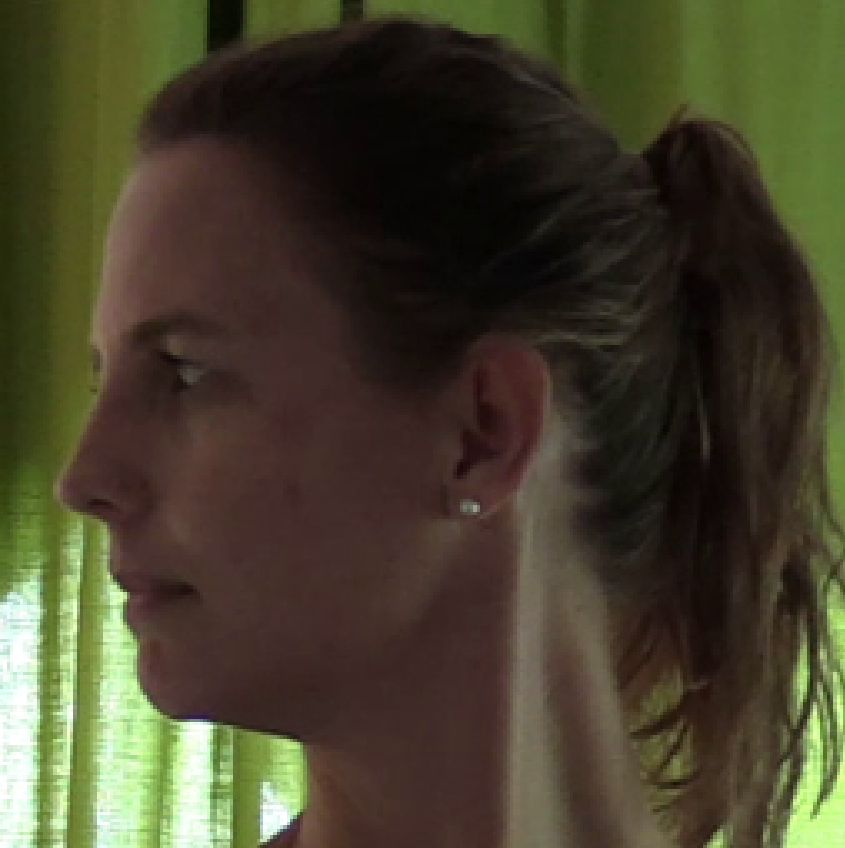}&
    \includegraphics[width=0.12\textwidth,height=0.12\textwidth]{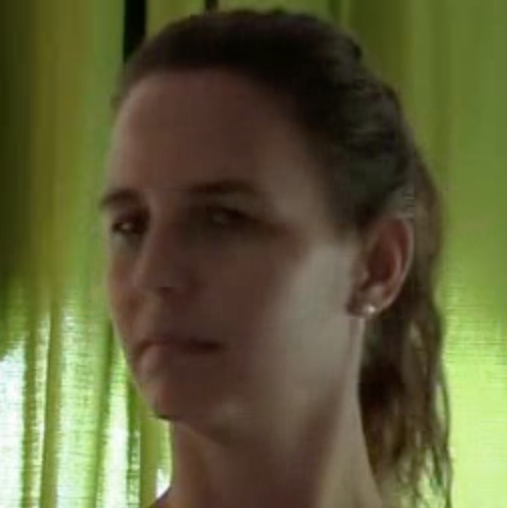}&
    \includegraphics[width=0.12\textwidth,height=0.12\textwidth]{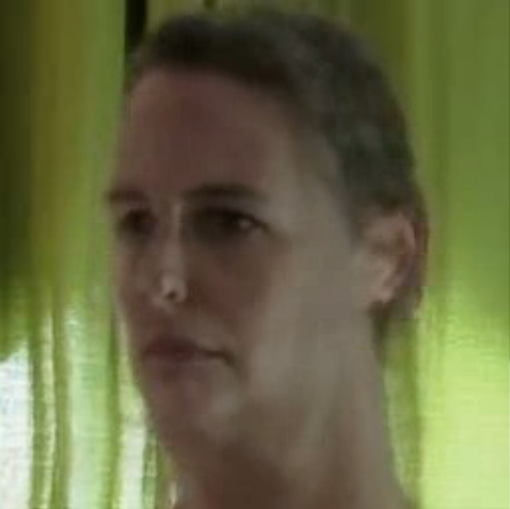}&
    \includegraphics[width=0.12\textwidth,height=0.12\textwidth]{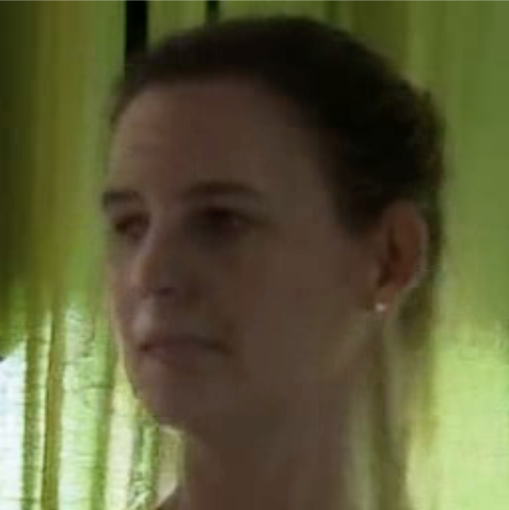}&
    \includegraphics[width=0.12\textwidth,height=0.12\textwidth]{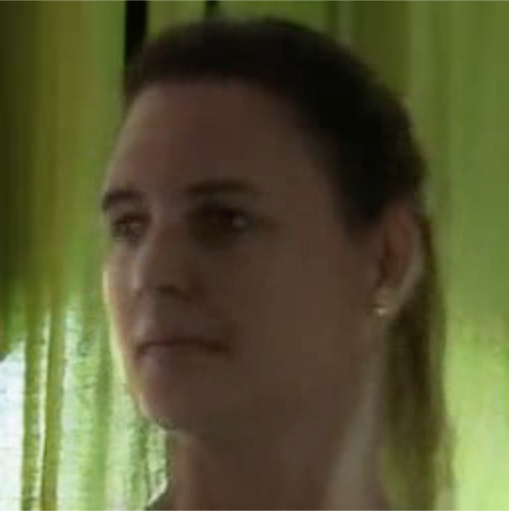}&
    \includegraphics[width=0.12\textwidth,height=0.12\textwidth]{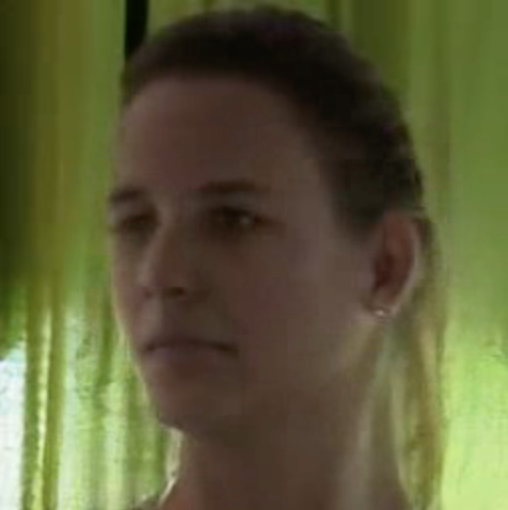}\\
    \end{tabular}
    \caption{Our dense Frontalizer variants reconstruct faces accurately where the FOM adv baseline struggles with occluded parts given its only source input $x_1$. The Frontalizer models also use the rightmost $x_2$ and leftmost $x_3$ frames.}
    \label{fig:densemodels}
\end{figure*}

\begin{table*}[ht]
    \centering
     \resizebox{1\textwidth}{!}{ 
    \begin{tabular}{lccccccccc}
    \toprule
    & \multicolumn{4}{c}{\textbf{DFDC-rotations}} & \multicolumn{4}{c}{\textbf{VoxCeleb}}\\
      & \small LPIPS $\downarrow$  & \small NME $\downarrow$ & \small CSIM $\uparrow$ & Expr $\uparrow$ & \small LPIPS $\downarrow$  & \small NME $\downarrow$ & \small CSIM $\uparrow$ & Expr $\uparrow$\\
    \midrule
    FOM adv & 0.136{\tiny $\pm$0.004}&1.108{\tiny $\pm$0.176}&0.784{\tiny $\pm$0.008}& 0.854{\tiny $\pm$0.008} & {\bf 0.118}{\tiny $\pm$0.003}&{\bf 0.247}{\tiny $\pm$0.007}&{\bf0.826}{\tiny $\pm$0.008}&	{\bf 0.915}{\tiny $\pm$0.003}\\
          Ours (Frontalizer sup) &0.132 {\tiny $\pm$0.002}&	0.496{\tiny $\pm$0.034}&0.793{\tiny $\pm$	0.005}&	0.879{\tiny $\pm$0.005} & 0.171{\tiny $\pm$0.004}&0.269{\tiny $\pm$0.004}&0.789{\tiny $\pm$0.005}&	0.897{\tiny $\pm$0.004}\\ % 20211022/facegen.QyDPsbxnoa bs 40
          Ours (Frontalizer sup+unsup) & {\bf 0.113}{\tiny $\pm$0.002}&0.469{\tiny $\pm$0.029}&{\bf0.799}{\tiny $\pm$0.005}&0.881{\tiny $\pm$0.005} & 0.151{\tiny $\pm$0.003}&0.255{\tiny $\pm$0.004}&0.791{\tiny $\pm$0.005}&	0.896{\tiny $\pm$0.004}\\ % 20211031/facegen.bjMuZfrvUP bs 30
          Ours (Frontalizer sup+unsup + free expr) & {\bf 0.113}	{\tiny $\pm$0.002}&	{\bf 0.467}{\tiny $\pm$	0.030}&	0.789{\tiny $\pm$0.005}&{\bf 0.897}{\tiny $\pm$	0.004} & 0.140{\tiny $\pm$0.003}&0.256{\tiny {$\pm$0.003}}&0.781{\tiny $\pm$0.005}&	0.910{\tiny $\pm$0.004}\\ %20211025/facegen.UnFlZUiFiq bs 30
          \midrule
          \ccc{Ours (Frontalizer sup+unsup)} {\bf new sampling} & {\bf 0.088} {\tiny $\pm$0.002}&	{\bf 0.425}{\tiny $\pm$0.027}&{\bf 0.850} {\tiny $\pm$	0.004}&	{\bf 0.902}{\tiny $\pm$0.004} & {\bf 0.116}{\tiny $\pm$0.002}&{\bf 0.237}{\tiny $\pm$0.002}&{\bf 0.837}{\tiny $\pm$0.002}& {\bf	0.915}{\tiny $\pm$0.002}\\
        \bottomrule 
    \end{tabular}}
    \caption{ Quality metrics of dense models. All models were trained on DFDC using adversarial training. The baseline FOM setup compares favorably on the VoxCeleb dataset, which present only little rotations, as shown in Fig.~\ref{fig:angle-distribution}. However our models compare favorably when large head angle rotations occur, as in the DFDC-rotations set.
    }
    \label{tab:QualityDM}
\end{table*}

\paragraph{Datasets.}

We evaluate our models on two datasets to ease comparisons with previous approaches: VoxCeleb \cite{Nagrani17vox} and DFDC \cite{dolhansky2019deepfake}. As shown in Fig.~\ref{fig:angle-distribution}, VoxCeleb contains only little rotations on yaw angle; 
therefore we extract from DFDC a validation set of 157 videos, comprising 90 to 350 frames each, filtered to contain a maximum of head movements ($\max_{\yaws} - \min_{\yaws} > 1.5$ rad), and call it ``DFDC-rotations". We also use the DFDC-50 diverse set shared by \cite{oquab2020low}.  

\paragraph{Metrics.}
We provide different measures to assess image generation:
The LPIPS \cite{zhang2018unreasonable} is a perceptual metric rating the overall quality of generations by comparing their deep features using a pre-trained network on image classification. 
Facial landmarks displacements are assessed by the NME as in \cite{bulat2017far}. To evaluate identity preservation, we rely on the CSIM metric, computing differences between Arcface features \cite{deng2019arcface}. Finally, to assess facial expressions preservation that can not be computed from landmarks only, we compute an $\ell_1$ norm of the difference of prediction of sixteen facial action units using a pre-trained network~\cite{toisoul2021estimation} to perform such classification. We denote this new metric as Expr.     

\paragraph{Frame selection setup}
At inference time, we select one source frame having the smallest yaw angle (frontal view) over the video, and ensure that eyes are open. For the Frontalizer models, we select two additional frames where the yaw angles are the extrema reached in the video. Examples of selected source frames are shown in Figure \ref{fig:densemodels}.

We demonstrate the quality improvements brought by our models on dense architectures in Section~\ref{sec:dense}, before presenting results on mobile architecture in Section~\ref{sec:mobile}. Finally, we quantify bandwidth requirements and compare to the previous on device state-of-the-art \cite{oquab2020low} in Section~\ref{sec:band}.

\subsection{Dense models}
\label{sec:dense}

We first provide qualitative results in Figure~\ref{fig:densemodels} that compares generations of dense models: The First Order Model (FOM adv), our result using a Frontalizer module with only supervised landmarks, using additional self-supervised landmarks, and using additional free expression conditioning. The progresses brought by the frontalization module, with respect to the FOM baseline are very significant. Indeed, the FOM baseline,  having only access to a source image with frontal pose, can not recover occluded face parts accurately. By design, our Frontalizer variants, thanks to their three source based embedding, infer realistic results. Among the Frontalizer variants, differences are less visible, however we note a better head shape preservation using the Sup+Unsup landmark setting, and a better background preservation using the free expression code. More examples are provided in Appendix.    

Table \ref{tab:QualityDM} provides quality metrics for the FOM model and our Frontalizer variants on DFDC-rotations and Voxceleb. \ccc{The frontalizer greatly outperforms FOM on DFDC-rotations.} As mentioned previously, the vast majority of VoxCeleb videos do not contain head rotations, and when they do, these are limited to low yaw angles. 
\ccc{In this context, it is not surprising that FOM reaches comparable performance. Using our novel sampling strategy, our approach manages to perform on par with FOM on VoxCeleb, even slightly  improving it.}  

We highlight that on mobile (cf. Table~\ref{tab:QualityMobile}), our models greatly outperform FOM on all datasets. The average NME of all approaches is in the range 0.247 - 0.256, indicating the degree of easiness of VoxCeleb for face animation. On DFDC-rotations, the diversity of head poses to render leads to a very high NME of 1.11 for FOM, and is much more reasonable for Frontalizer models, reaching 0.467 for the best model conditioned on expressions. The CSIM and LPIPS are further improved using additional \Unsup~landmarks, and the Expr metric using free expression conditioning.

\paragraph{Ablation study for the expression module.}

\begin{figure}[ht]
    \centering
      \setlength{\tabcolsep}{1pt}
    \begin{tabular}{ccccc}
    Driving & FOM  & Dec cond. & Free cond.\\
     \includegraphics[width=0.24\textwidth]{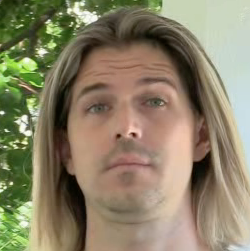} &
     \includegraphics[width=0.24\textwidth]{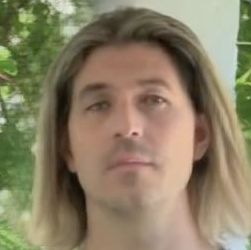} &
     \includegraphics[width=0.24\textwidth]{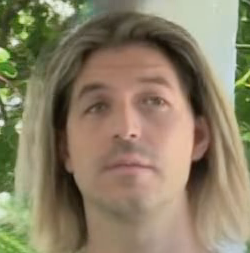} &
     \includegraphics[width=0.24\textwidth]{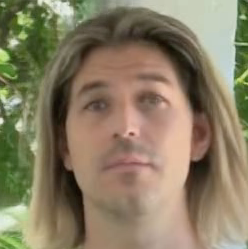} 
    \end{tabular}
    \caption{Ablation study of expression conditioning. Free decoder conditioning keeps the wrinkles while limiting head distortions.}
    \label{fig:wrinkles}
\end{figure}

\begin{figure}[ht]
    \centering
     \includegraphics[width=0.32\textwidth]{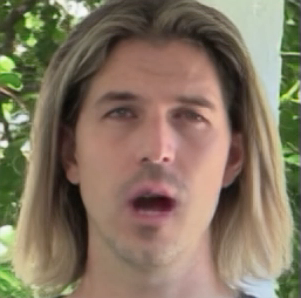} \includegraphics[width=0.32\textwidth]{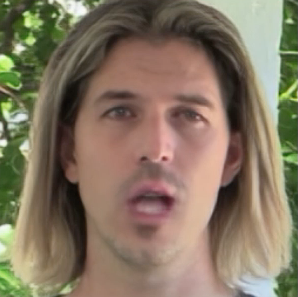}
    \includegraphics[width=0.32\textwidth]{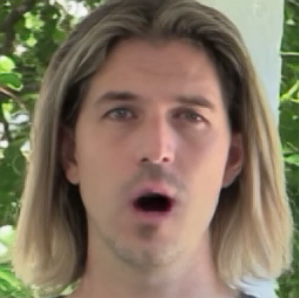}   
    \caption{Variation in the free expression latent space (a): using original driving code $e$. (b) Setting $e_7$ to 1; (c) Setting $e_6$ to 0.}
    \label{fig:free_expr}
\end{figure}

We present in Table \ref{tab:QualityAblationExpression} different variants of the expression conditioning we tried on top of the FOM baseline. 
First, we check that the addition of the Facial action unit loss alone does not help. 
Second, we compare expression conditioning by feeding ``oracle'' expression code in either the Dense motion network or in the Decoder. Adding these conditioning leads to slightly worse LPIPS but better expression metrics, with a larger improvement in expression brought by decoder conditioning. The boost brought by the light supervised code conditioning is small. Finally,  the free light conditioning strategy is cheap to allow for mobile inference, does not increase the LPIPS significantly, and brings an improvement to the expression metric, justifying selecting this conditioning strategy for our experiments. 

 \begin{figure}[ht]
    \centering 
    Sources \\
     \setlength{\tabcolsep}{1pt}
      \begin{tabular}{ccccccccc}
       \includegraphics[width=0.1\textwidth,height=0.1\textwidth]{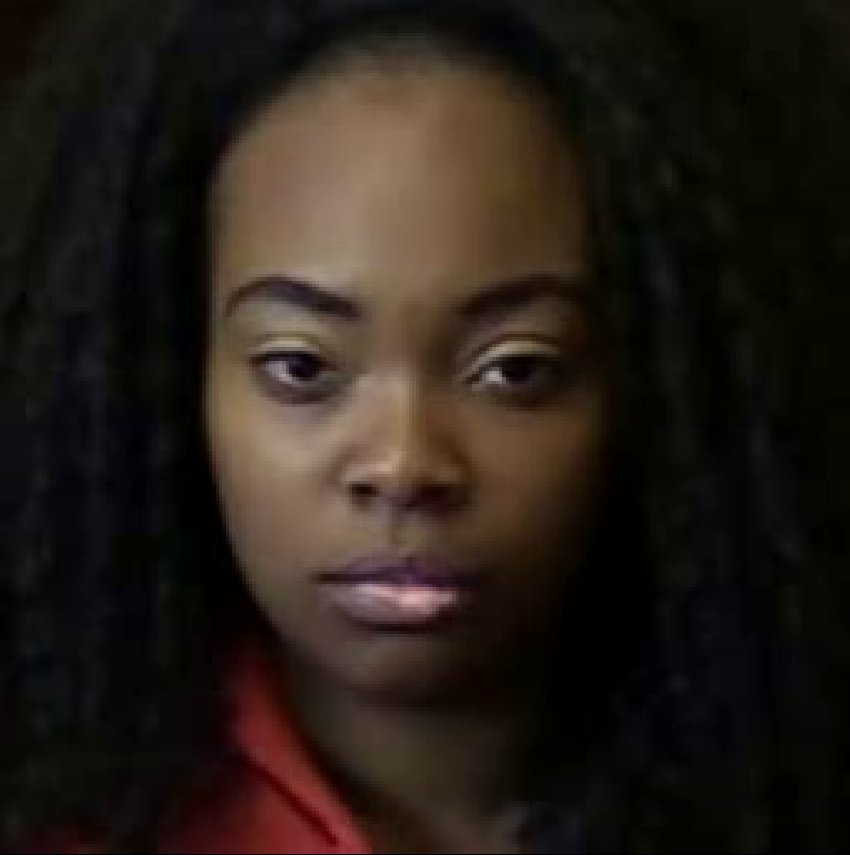}& 
       \includegraphics[width=0.1\textwidth,height=0.1\textwidth]{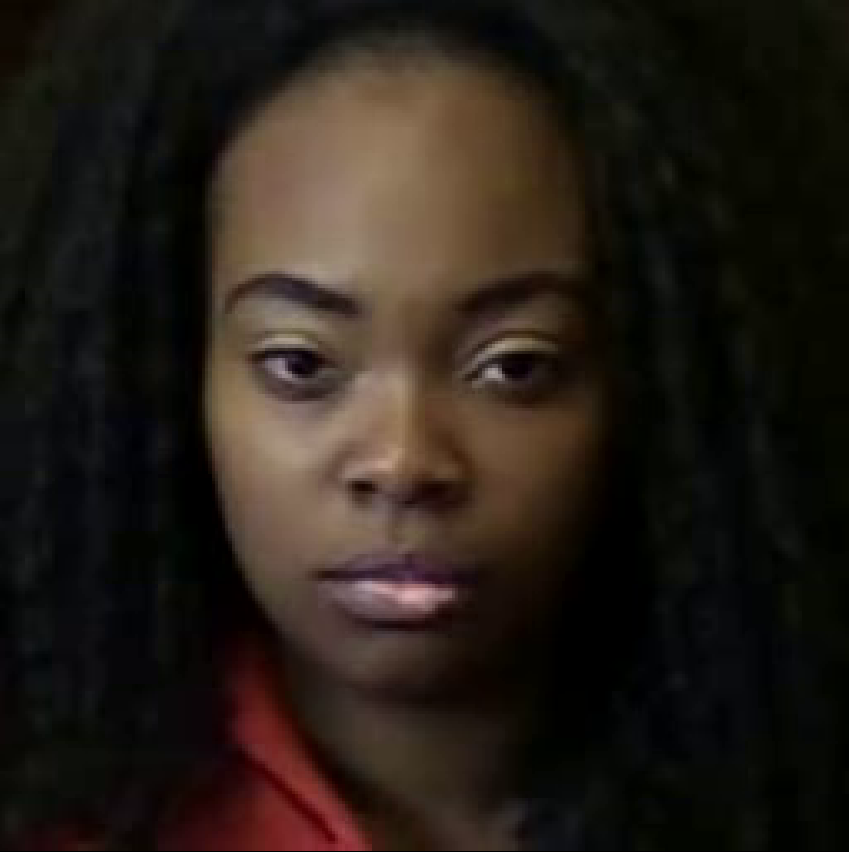}& 
       \includegraphics[width=0.1\textwidth,height=0.1\textwidth]{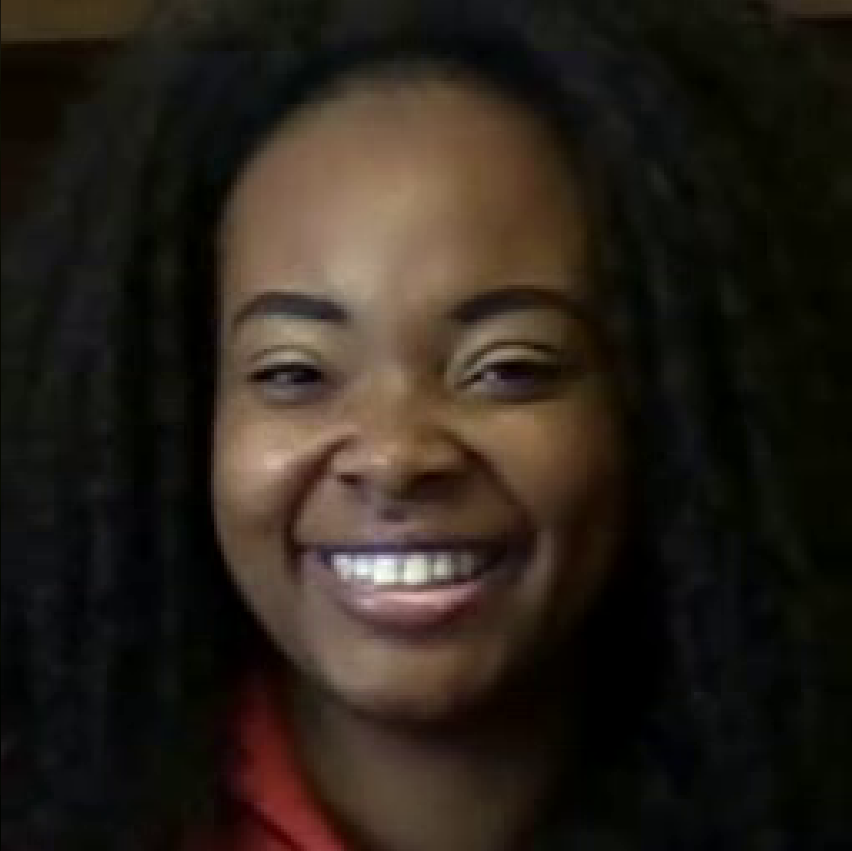}&
         \includegraphics[width=0.1\textwidth,height=0.1\textwidth]{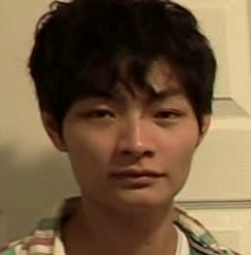}& 
       \includegraphics[width=0.1\textwidth,height=0.1\textwidth]{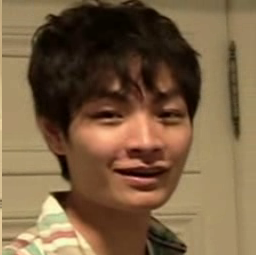}& 
       \includegraphics[width=0.1\textwidth,height=0.1\textwidth]{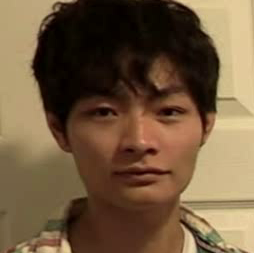}&
         \includegraphics[width=0.1\textwidth,height=0.1\textwidth]{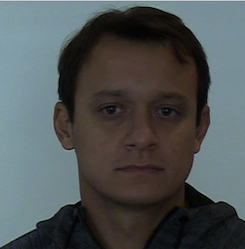}& 
       \includegraphics[width=0.1\textwidth,height=0.1\textwidth]{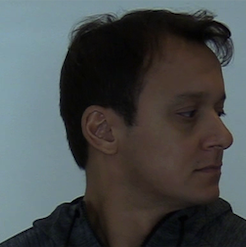}& 
       \includegraphics[width=0.1\textwidth,height=0.1\textwidth]{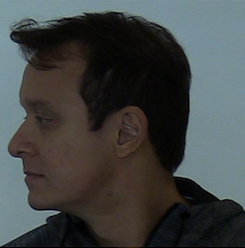}
            \end{tabular}
                 \setlength{\tabcolsep}{5pt}
    \begin{tabular}{ccc}
 & Driving &  \\
    \includegraphics[width=0.27\textwidth,height=0.27\textwidth]{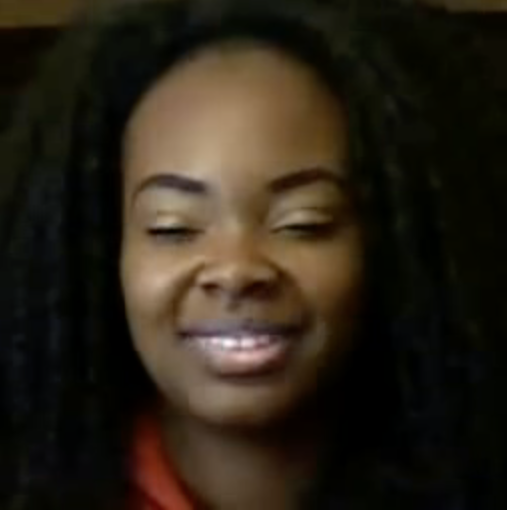}& 
    \includegraphics[width=0.27\textwidth,height=0.27\textwidth]{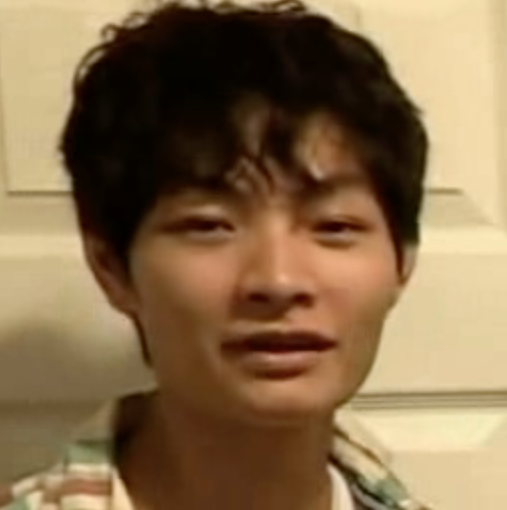}&
    \includegraphics[width=0.27\textwidth,height=0.27\textwidth]{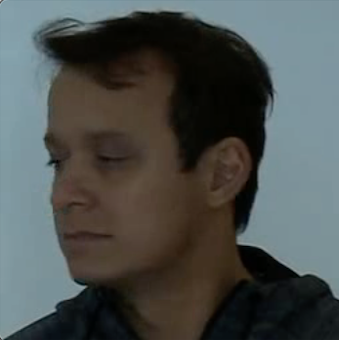}\\
    & FOM adv & \\
    \includegraphics[width=0.27\textwidth,height=0.27\textwidth]{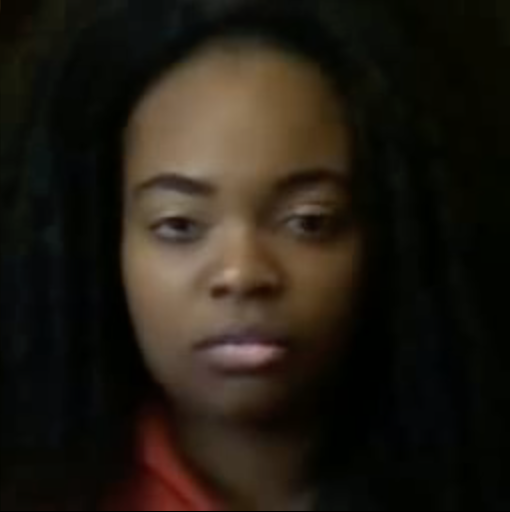}& 
    \includegraphics[width=0.27\textwidth,height=0.27\textwidth]{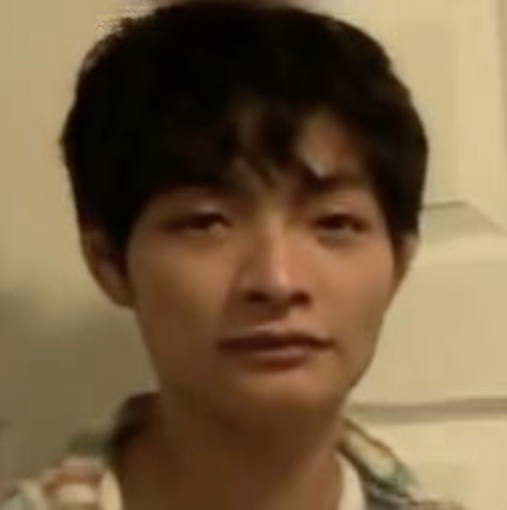}&
    \includegraphics[width=0.27\textwidth,height=0.27\textwidth]{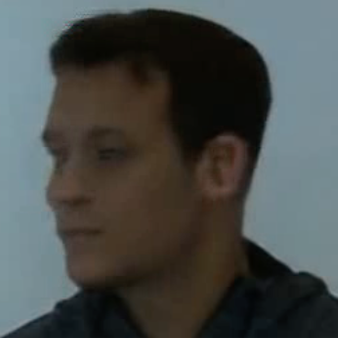}\\
    \end{tabular}
      Frontalizer \Sup+\Unsup \\
     \begin{tabular}{ccc}
     \includegraphics[width=0.27\textwidth,height=0.27\textwidth]{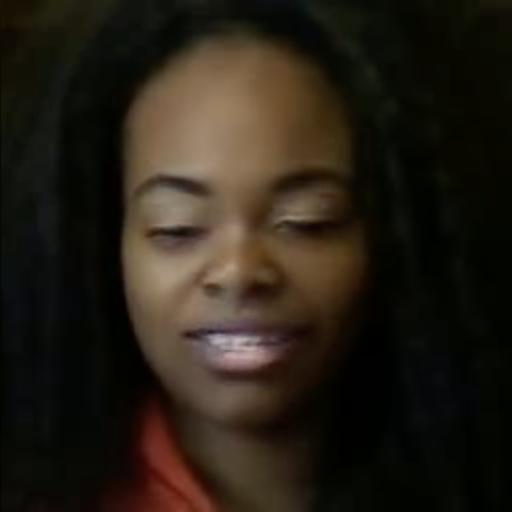}& 
    \includegraphics[width=0.27\textwidth,height=0.27\textwidth]{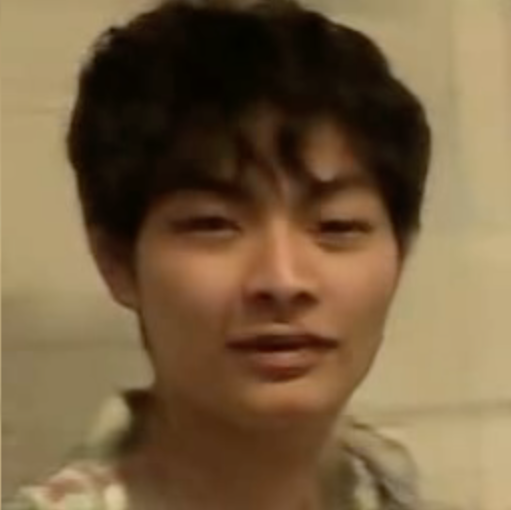}&
    \includegraphics[width=0.27\textwidth,height=0.27\textwidth]{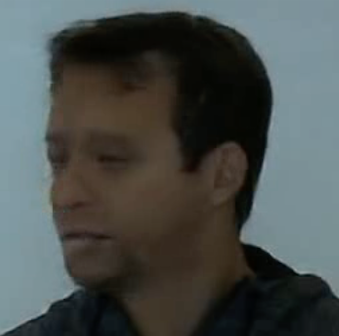}\\
    & HEVC (8 kbps) & \\
     \includegraphics[width=0.27\textwidth,height=0.27\textwidth]{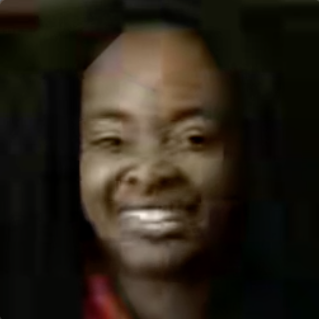}& 
    \includegraphics[width=0.27\textwidth,height=0.27\textwidth]{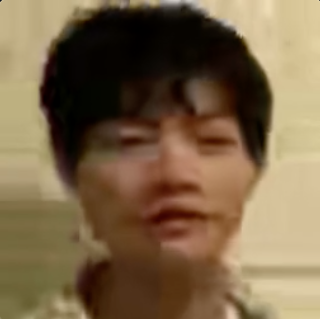}&
    \includegraphics[width=0.27\textwidth,height=0.27\textwidth]{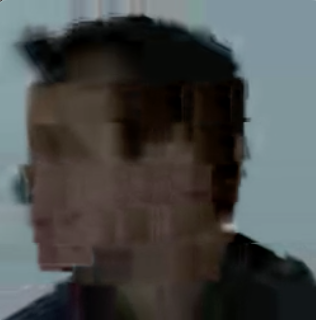}
      \end{tabular}
    \caption{Our mobile Frontalizer models outperform FOM adv, requiring less than 8 kbps, and visibly improve over HEVC.}
    \label{fig:mobile}
\end{figure}

Qualitatively, as shown in Figure \ref{fig:wrinkles}, our free conditioning helps to recover a good part of the wrinkles of the driving frames, while reducing the presence of hair artifacts in the background. 
It is possible to navigate in the free latent space and observe the effect of forcing different values for conditioning code. In Figure~\ref{fig:free_expr}, setting some values of the code to zero or one leads to various mouth interior appearances.

\paragraph{Quantitative evaluation of dense models.}

\begin{table}[ht]
    \centering
     \resizebox{0.9\textwidth}{!}{ 
    \begin{tabular}{lccc}
    \toprule
      &  LPIPS $\downarrow$  & CSIM $\uparrow$ & Expr $\uparrow$\\
    \midrule
           FOM & 0.115  & 0.841 & 0.934\\ %bs 32 dfdc FOM 20211015/facegen.chBLbEfQIl
           FOM + FAU loss & 0.119 & 0.822 & 0.926\\ % qas/20211112/facegen.adkTzVxCDe
           FOM + DM expr cond. & 0.117  & 0.833 & 0.951\\ %20211021/facegen.ekDEvswgrg 
           FOM + Dec expr cond.  & 0.118 & 0.834 & 0.964 \\ %20211021/facegen.SdSSAOmoZx
           FOM + light Dec expr cond.  & 0.118  & 0.834&  0.939\\ %DM:  20211021/facegen.IyWcfGywsT
           FOM + light Dec free cond.  &  0.116  & 0.834 & 0.949\\ % 20211022/facegen.avzyAqyYyd
        \bottomrule 
    \end{tabular}}
    \caption{Ablation study for expression conditioning on DFDC-50. The free Decoder conditioning strategy is the best compromise.}
    \label{tab:QualityAblationExpression}
\end{table}

\begin{table*}[h]
    \centering
    \resizebox{1\textwidth}{!}{ 
    \begin{tabular}{lccccccccc}
    \toprule
    & \multicolumn{4}{c}{\textbf{DFDC-rotations}} & \multicolumn{4}{c}{\textbf{VoxCeleb}}\\
      & \small LPIPS $\downarrow$  & \small NME $\downarrow$ & \small CSIM $\uparrow$ & Expr $\uparrow$ & \small LPIPS $\downarrow$  & \small NME $\downarrow$ & \small CSIM $\uparrow$ & Expr $\uparrow$\\
    \midrule
          FOM-Mobile & 0.187 {\tiny $\pm$0.004}& 1.122 {\tiny $\pm$0.105}& { 0.722} {\tiny $\pm$0.005}& 0.790 {\tiny $\pm$0.006}& 0.197 {\tiny $\pm$0.002}& 0.519 {\tiny $\pm$0.007}& 0.717 {\tiny $\pm$0.004}& 0.837 {\tiny $\pm$0.003}\\ % 20211022/facegen.QyDPsbxnoa bs 40
          Ours (sup) & 0.172 {\tiny $\pm$0.003}& 0.632 {\tiny $\pm$0.062}& 0.710 {\tiny $\pm$0.006}& {\bf 0.866} {\tiny $\pm$0.005}& 0.184 {\tiny $\pm$0.002}& 0.288 {\tiny $\pm$0.003}& 0.736 {\tiny $\pm$0.003}& 0.890 {\tiny $\pm$0.002}\\
          Ours (sup+unsup) & 0.148 {\tiny $\pm$0.003}& { 0.585} {\tiny $\pm$0.044}& { 0.723} {\tiny $\pm$0.006}& { \bf 0.865} {\tiny $\pm$0.005}& { \bf 0.165} {\tiny $\pm$0.002}& { 0.275} {\tiny $\pm$0.003}& { 0.747} {\tiny $\pm$0.003}& {\bf 0.893} {\tiny $\pm$0.002}\\
          \midrule
  \ccc{Ours (sup+unsup)*} & {\bf 0.142} {\tiny $\pm$0.003}& {\bf 0.493} {\tiny $\pm$0.035}& {\bf 0.793} {\tiny $\pm$0.005}& { 0.851} {\tiny $\pm$0.005}& { 0.175} {\tiny $\pm$0.002}& {\bf 0.250} {\tiny $\pm$0.003}& {\bf 0.799} {\tiny $\pm$0.003}& {0.883} {\tiny $\pm$0.002}\\
        \bottomrule 
    \end{tabular}}
   
    \caption{ Quality metrics of mobile models. *: model with new sampling and SPADE based decoder as in \cite{oquab2020low}. The new sampling strategy combined to SPADE layers improves results, particularly in terms of identity preservation.
    }
    \label{tab:QualityMobile}
\end{table*}

To ensure that the findings given by automated metrics correspond to human perception, we conducted a human annotation of the quality of reconstructed videos using the four models of Table \ref{tab:QualityDM}, showing videos from DFDC-rotations and DFDC-50 to 15 annotators. $12\%$ of FOM videos were preferred, $20\%$ of Frontalizer Sup,  $46\%$ of Frontalizer Sup+Unsup, and $22\%$ of Frontalizer S+U with expression free conditioning. 
% 0= 11.5942029 1= 19.80676329 2 = 46.37681159 3 = 22.22222222

\subsection{Mobile models}
\label{sec:mobile}

In this section, we provide results obtained with mobile models. We replace the dense networks with their mobile counterparts, using MobileNetV2 building blocks, and the same network architectures for our approach and for the FOM-Mobile baseline, based on \cite{Siarohin_2019_NeurIPS} and using ten unsupervised keypoints with their Jacobians. We also reduce the size of the bottleneck feature from $64\times64$ to $32\times32$ to save on computation and ensure a 30 FPS frame rate on iPhone 8 for all models. We report our results in Table~\ref{tab:QualityMobile}. We observe that for both VoxCeleb and DFDC, in the mobile computation regime, our models using the Frontalizer compare favorably to the baseline FOM-Mobile, for image quality (LPIPS) and landmark fidelity (NME). Expression did not bring improvement in the metrics as detailed in the Appendix. Qualitative comparisons appear in Figure~\ref{fig:mobile}. The results show that in low-compute and low-bandwidth regime, it is beneficial to add extra facial landmarks and rely on multiple frames: we observe that eye blinking and large mouth opening are skipped by the FOM mobile baseline.

\paragraph{Influence of SPADE in the decoder.}

\ccc{Finally, we train a model combining our Frontalizer approach using the new sampling strategy and SPADE layers in the decoder \cite{oquab2020low}. The obtained  gains are significant both numerically (cf. Table \ref{tab:QualityMobile}) and qualitatively for rendering pupils and teeth. For training this final model, we disabled adversarial training, the learning rate scheduler and let the model train for 400K iterations. The bandwidth of this model is 10.7 kbps at 15 frames per second. }

\paragraph{Inference time.}

Our mobile models (including the landmark tracker) run at 30 FPS on an iPhone 8 using CoreML to access the GPU chip.
In our measurements, for mobile models capped at 15 FPS, the battery usage is around $0.2\%$ / min on recent phones (e.g. 2019 iPhone 11 with a specialized Neural Engine chip), and around $0.7\%$ / min on older phones such as the iPhone 8.

\subsection{Frame compression}
\label{sec:band}

To reconstruct the current frame on the receiver side, we need the source frame(s), and the driving frame's necessary information. We assume that source frames are sent ahead of time and focus our bandwidth estimate on the driving frame information. 
A driving sample can be described by a list of floating-point numbers: the average keypoint position, 
the position of the keypoints relative to this average position, the Jacobians and the expression codes. 
Rather than sending these values directly, we instead send deltas compared to the previous frame's values. 
We furthermore quantize \Unsup~keypoint coordinates on 12-bits, \Sup~keypoints on 8 bits, the $4\times \nkp$ Jacobians on 16 bits in $[-15, 15]$, and the $\nexpr$ expression codes on 10 bits in $[-1, 1]$.
We then adopt arithmetic coding \cite{rissanen1979arithmetic} to compress the data. Arithmetic coding is a lossless entropy coding algorithm, using a prior distribution over the values to encode frequent values using fewer bits. 
We compute this distribution on the first half of the DFDC val. set, and report compressed sizes on the second half for all models in Table~\ref{tab:bandwidths}.

\begin{table}[h]
    \centering
    \begin{tabular}{lccc}
    \toprule
     Model & Bits per frame & Bdth @15fps \\
      \midrule
      FOM {\small(\Unsup~+Jac.)} & 426 & 6.4 kbps \\ %483 \\
     Ours {\small(\Sup)} & 264 & 4.0 kbps \\ %362\\
     Ours {\small(\Sup+\Unsup)} & 418 & 6.3 kbps\\%\td{x}\\
     Ours {\small(\Sup+\Unsup+Expr)} & 433 & 6.5 kbps\\%\td{x}\\
    \bottomrule
    \end{tabular}
    \caption{Frame-size for different models, compressed with arithmetic encoding, and measured on 15 fps videos of the test set.}
    \label{tab:bandwidths}
\end{table}

\ccc{
\subsection{Additional comparisons}}

\begin{figure}[ht]
  \centering
  \setlength{\tabcolsep}{1pt}
  \begin{tabular}{cccc}
  {\scriptsize FOM-mult frame $t$ } & {\scriptsize FOM-mult $t+1$} & {\scriptsize Ours frame $t$} & {\scriptsize Ours $t+1$}\\
  \includegraphics[height=0.23\linewidth]{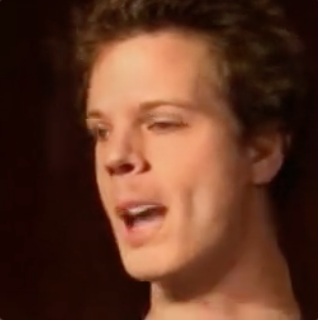}& \includegraphics[height=0.23\linewidth]{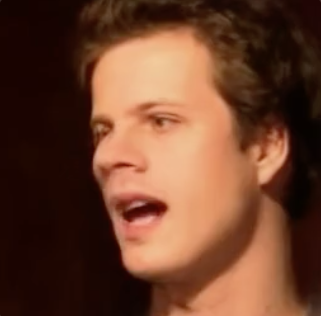}& 
  \includegraphics[height=0.23\linewidth]{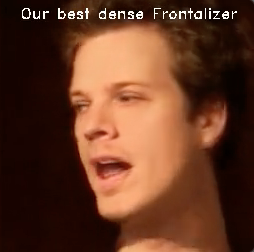}&
  \includegraphics[height=0.23\linewidth]{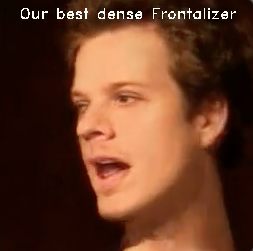}
  \end{tabular}
  \caption{The multiple reference frame FOM baseline still suffers from head deformations and results in non-smooth transitions when dynamically switching the source frame.}
   \label{fig:mult}
\end{figure}

\ccc{We provide numerical comparisons to two competing approaches NTH \cite{zakharov2019few}, Hybrid SPADE \cite{oquab2020low}, and a third baseline dubbed ``FOM mult'' in Table \ref{table}.
The FOM mult baseline consists in dynamically selecting the reference frame in which landmarks are the closest to the driving one, and apply the FOM approach to reconstruct the result. 
We include a qualitative comparison to this baseline in Fig. \ref{fig:mult} and quantitative results in Table \ref{table}. Even if their is a slight drop in LPIPS on dense models, the flickering it induces when the head turns makes it unusable in practice.} 

\begin{table}[h]
    \centering
    \resizebox{1.0\linewidth}{!}{
    \begin{tabular}{lccc}
    \toprule
      &  LPIPS $\downarrow$ & NME $\downarrow$ & CSIM $\uparrow$ \\
    \midrule
           FOM \cite{Siarohin_2019_NeurIPS} & 0.116 \improv{12} & 0.375 \improv{31} & 0.817 \improv{6} \\ % 0.116	0.084	0.375	0.817	0.920
           FOM mult & {\bf 0.092} \down{11} & 0.318 \improv{19} & 0.851 \improv{2}\\
           NTH \cite{zakharov2019few} & 0.147 \improv{31} & 0.288 \improv{10}& 0.785 \improv{10}\\ 
           Ours dense & 0.102\spacefill{xx} & {\bf 0.259}\spacefill{xx} & {\bf 0.864}\spacefill{xx}\\ %0.102	0.075	0.259	0.864	0.935
    \midrule 
            Hybrid-SPADE [11] & 0.166 \improv{10} & 0.312 \improv{9} & 0.801 \improv{2}\\
           Ours mobile & {\bf 0.149}\spacefill{xx} & {\bf 0.285}\spacefill{xx} & {\bf 0.821}\spacefill{xx}\\ % 0.149	0.113	0.285	0.821	0.908
        \bottomrule 
    \end{tabular}}
    \caption{Comparison of dense and mobile methods on DFDC-50. The FOM with multiple reference baseline, reported for reference, is not temporally consistent, thus inapplicable (See Fig.~\ref{fig:mult}).}
    \label{table}
\end{table}

\begin{figure}[ht]
  \centering
    \setlength{\tabcolsep}{1pt}
  \begin{tabular}{ccc}
  Driving frame & NTH result [21] & Our result\\
  \includegraphics[height=0.27\linewidth]{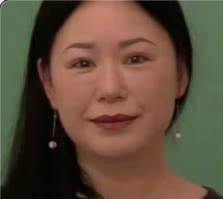}& \includegraphics[height=0.27\linewidth]{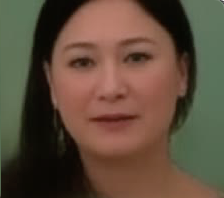}& 
  \includegraphics[height=0.27\linewidth]{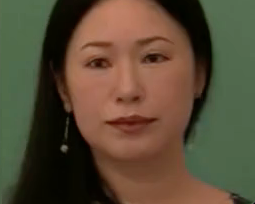} 
  \end{tabular}
   \caption{Our Frontalizer preserves identity better than Neural Talking Heads \cite{zakharov2019few}.}
   \label{fig:NTH}
\end{figure}

\ccc{ The Neural Talking Head (NTH) approach \cite{zakharov2019few} suffers from a loss of identity (low CSIM), as illustrated in Fig.~\ref{fig:NTH}.} 

\begin{figure}[ht]
    \centering 
    \setlength{\tabcolsep}{1pt}
     \resizebox{0.9\textwidth}{!}{ 
   \begin{tabular}{ccc}
   Driving & Result \cite{oquab2020low}  & Our result\\
    \includegraphics[width=0.32\textwidth, height=0.32\textwidth]{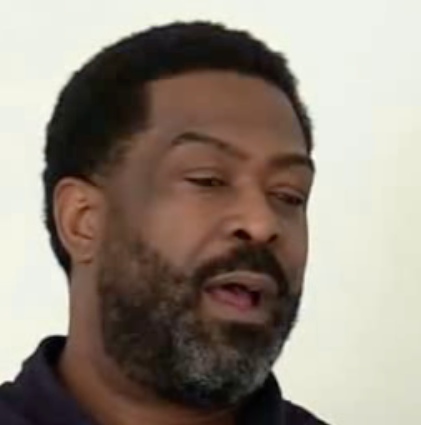} &
     \includegraphics[width=0.32\textwidth,height=0.32\textwidth]{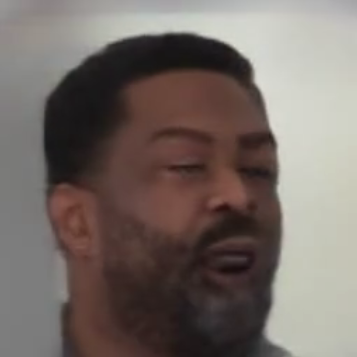} &
      \includegraphics[width=0.32\textwidth,height=0.32\textwidth]{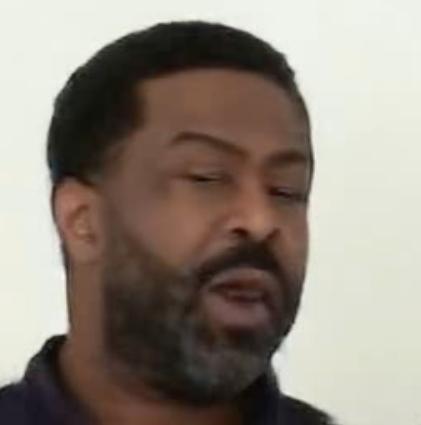} \\
       \includegraphics[width=0.32\textwidth,height=0.32\textwidth]{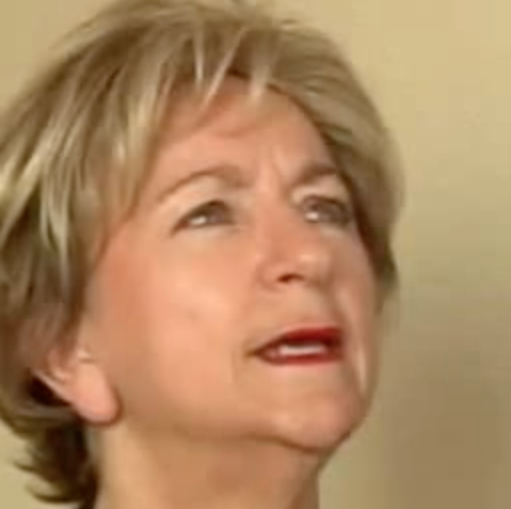} &
     \includegraphics[width=0.32\textwidth,height=0.32\textwidth]{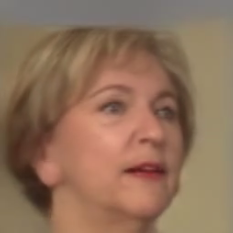} &
      \includegraphics[width=0.32\textwidth,height=0.32\textwidth]{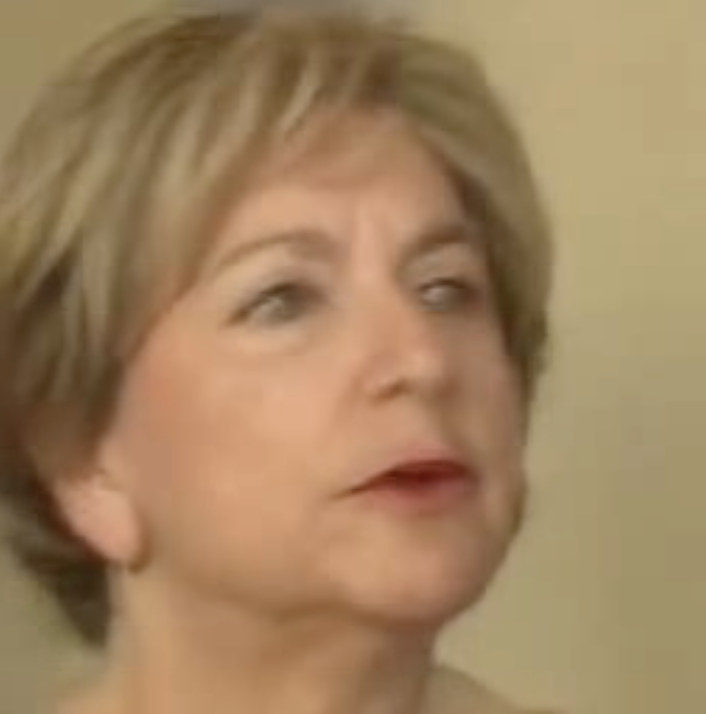} 
     \end{tabular}}
    \caption{Comparison to Hybrid Spade \cite{oquab2020low} on mobile.    
 }
    \label{fig:mobile_comp}
\end{figure}

Figure~\ref{fig:mobile_comp} provides a comparison of mobile results with the best model from \cite{oquab2020low}. Our results are sharper with a twice larger resolution of $256\times$ and preserve expressions better. \ccc{Most importantly, our approach can be combined with \cite{oquab2020low}, leading to state-of-the-art results on mobile, combining the strengths of the Frontalizer embedding with the refinement of hybrid SPADE layers.}

\section{Conclusion}

\paragraph{Contributions.}
We propose a solution to the profile views rendering problem for face animation by designing a frontalized embedding based on a few source images warped to a reference position in the feature space. The resulting multi-view embedding avoids relying on the model to infer unseen areas during inference. We show that our approach compares favorably to baselines, \ccc{especially} in the case of large head rotations, with mobile results outperforming the state-of-the-art by a large margin. We also propose using a lightweight conditioning approach to refine facial expressions improving dense animation approaches. 

\paragraph{Limitations.}

While the majority of examples are correctly rendered with our approaches, there remains some margin for improvements in mobile models for glasses, pupils, head occlusions.
Temporal consistency is dependent on the landmark extractor. In general we do not observe problems but slight flickering may happen in case of bad landmark detections.

\paragraph{Broader impacts.}

Deep facial image generation is a sensitive research topic because reenactment applications with bad intentions could be harmful. We will not publish the source code for this work, to reduce these risks. 
We hope that the good performance of our approach in low compute regime will help to unlock video chat in low bandwidth situations.

\subsubsection*{Acknowledgments}
We would like to thank Michael Chang, Patrick Labatut, Jeff LaFlam, Maja Pantic,  Thibault Peyronel, Albert Pumarola, Marcy Regalado for their various support to our work. 
%%%%%%%%%%%%%%%%%%%%%%%%%

{\small
\bibliographystyle{ieee_fullname}
\bibliography{fercvpr}
}

\ccc{
\section{Appendix}}

\subsection{Architectures and training details}

\subsubsection{Frontalizer}

The Frontalizer module contains an hourglass network and a set of keypoints defining the reference position $p_r$. In the server (\textit{dense}) models, the architecture of the hourglass net is identical to the one of the Dense Motion network of \cite{Siarohin_2019_NeurIPS}. In this section we will focus on describing the mobile architectures.

\begin{figure*}[h]
    \centering
    \includegraphics[width=0.75\textwidth, page=1, trim=0 7.5cm 9cm 0, clip=true]{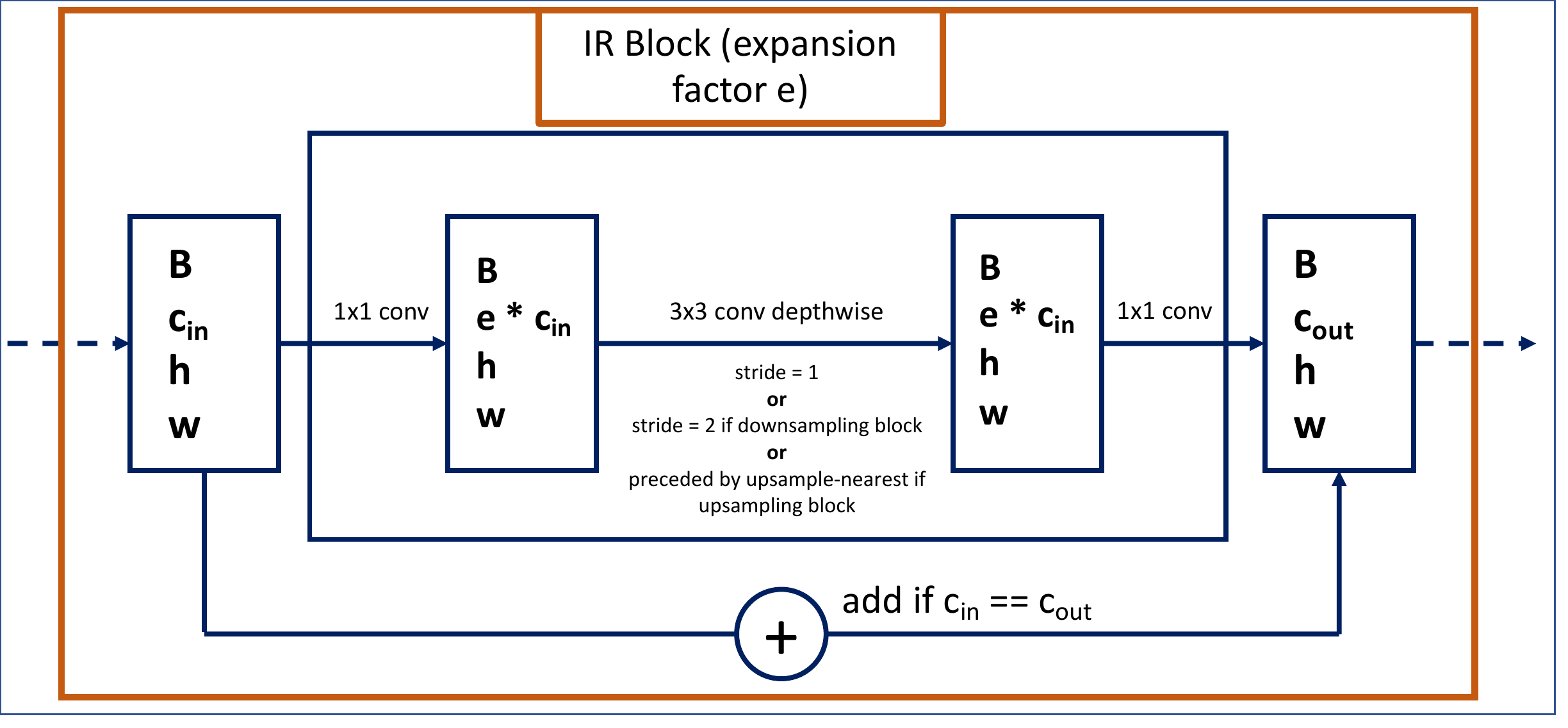} 
    \caption{MobileNetV2 Inverted Residual block with expansion factor $e$, for a number of input channels $c_{in}$ and output channels $c_{out}$.}
    \label{fig:supmat-irblock}
\end{figure*}

\begin{figure*}[h]
    \centering
    \includegraphics[width=0.95\textwidth, page=2, trim=0 5.0cm 2cm 0, clip=true]{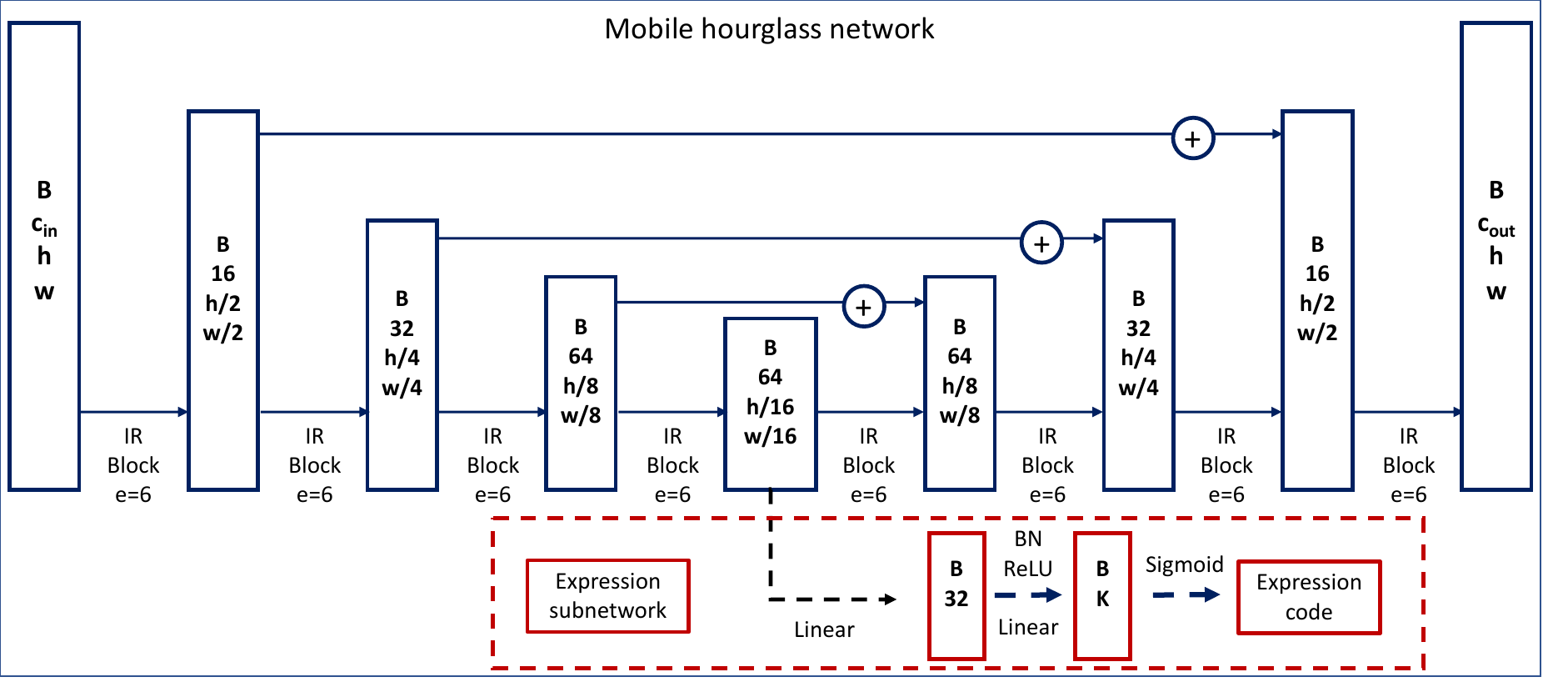} 
    \caption{Mobile hourglass architecture used in our mobile setups, for the Keypoint Detector, Dense Motion and Frontalizer networks. In order to extract the expression code, we add a small MLP network that uses the central feature of the Keypoint Detector network.}
    \label{fig:supmat-hourglass}
\end{figure*}

The base building block of our networks is the inverted residual block introduced in \cite{sandler2018mobilenetv2}, that we describe in Figure~\ref{fig:supmat-irblock}. We then describe the mobile architectures for the different networks: the Frontalizer, Dense Motion, Keypoint and expression networks. All of them use the hourglass architecture defined in Figure~\ref{fig:supmat-hourglass}.

 The number of source frames used for training is set to $\nsrc=2$ ($\nsrc$ can be higher during inference). We note $K$ the number of unsupervised keypoints when applicable, and we note $n_{kp}$ the total number of keypoints including the facial landmarks.

The Frontalizer network takes as input a tensor of size $\left(\nsrc, n_{kp}, 32, 32\right)$, where each map of size $32\times32$ describes a keypoint displacement as a difference of Gaussians. It outputs $\nsrc$ flow maps of size $\left(32, 32, 2\right)$ and $\nsrc$ confidence maps of size $\left(32, 32\right)$. The flow maps are then applied to the encoder feature for each of the $\nsrc$ images to obtain warped features, and the confidence maps are softmax-ed, then used for obtaining the weighted sum of the warped features, corresponding to the multi-view embedding. 

The dense motion network takes as input a tensor of size $\left(n_{kp}, 32, 32\right)$ describing the keypoints displacements, and outputs a one flow map of size $\left(32, 32, 2\right)$ and one confidence map of size $\left(32, 32\right)$, that are both applied to the multi-view embedding.

The (unsup) keypoint extractor takes as input an image tensor of size $\left(3, 64, 64\right)$ and outputs $K$ heatmaps of size $\left(64, 64\right)$. We apply a spatial softmax to these heatmaps then compute the center of mass of the resulting maps to output $2 \times K$ keypoint coordinates.

\begin{figure*}[h]
    \centering
    \includegraphics[width=0.75\textwidth, page=3, trim=0 7.5cm 14.5cm 0, clip=true]{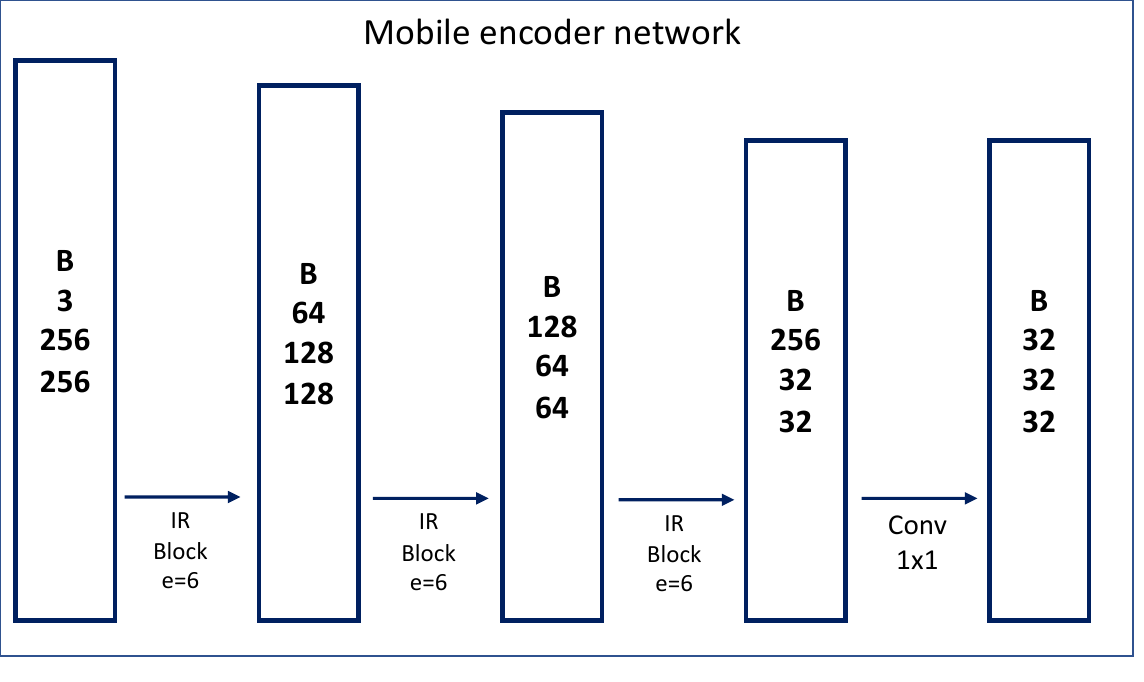} 
    \caption{Mobile encoder architecture used in our mobile setups}
    \label{fig:supmat-encoder}
\end{figure*}

\begin{figure*}[h]
    \centering
    \includegraphics[width=0.95\textwidth, page=4, trim=0 6.5cm 5cm 0, clip=true]{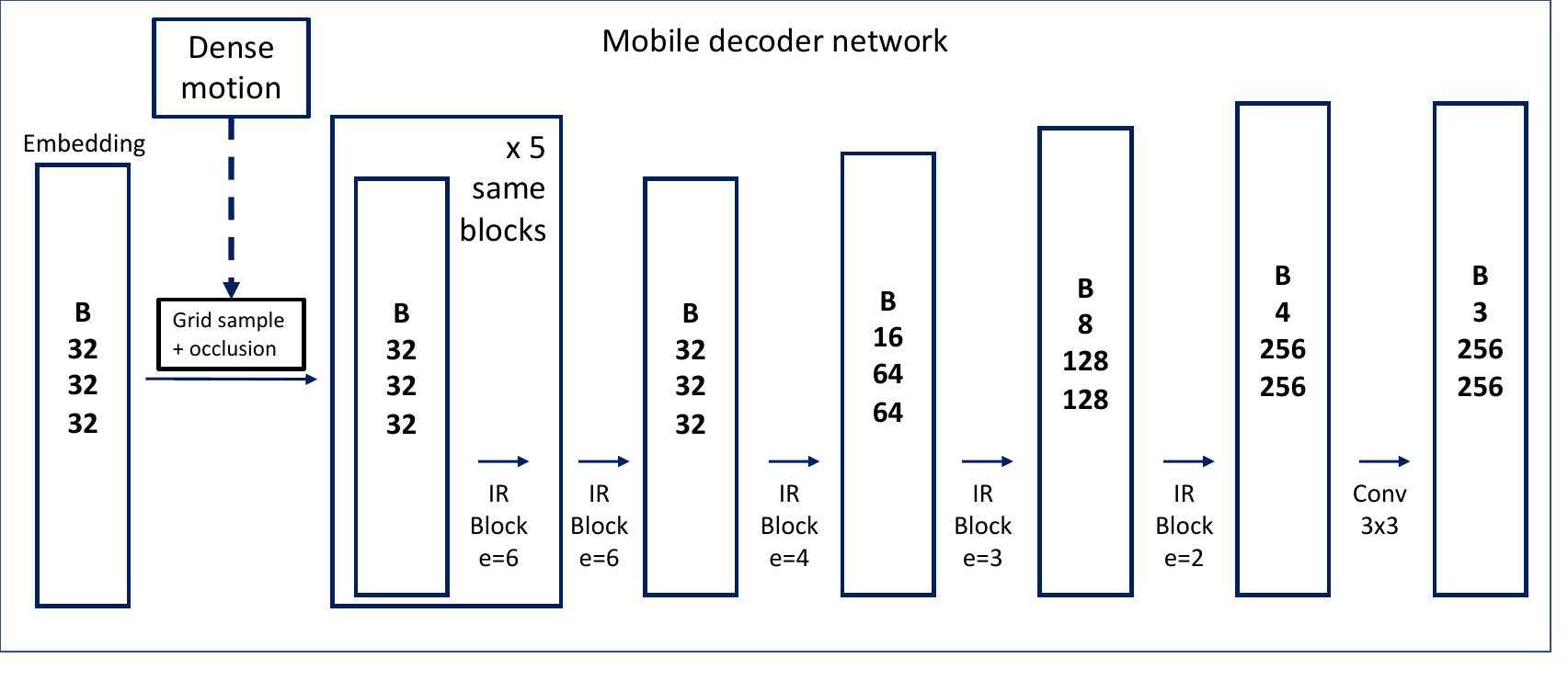} 
    \caption{Mobile decoder architecture used in our mobile setups}
    \label{fig:supmat-decoder}
\end{figure*}

The Encoder, described in Figure~\ref{fig:supmat-encoder}, takes as input an image tensor of size $\left(3, 256, 256\right)$ and outputs a feature map of size $\left(32, 32, 32\right)$. 
The Decoder, described in Figure~\ref{fig:supmat-decoder}, takes as input a feature map of size $\left(32, 32, 32\right)$ and outputs an image of size $\left(3, 256, 256\right)$.

\subsubsection{Dense expression module}

The MLP $g$ is a two-layer network taking in input a 4096 dimensional vector for dense models, or 512 dimensional for mobile models. It consists of a linear layer mapping to a 32 dimensional layer, a batch normalization layer, a ReLU, a linear layer mapping to a 16 dimensional vector and a Sigmoid.
For training, the weight $\gamma_E$ of the expression loss is set to 40.

\subsubsection{Training losses}

In addition to the multi-scale perceptual loss, and equivariance loss in settings employing Unsup keypoints, we employ a feature matching loss, and an adversarial loss. We use the same hyper-parameters as defined in \cite{Siarohin_2019_NeurIPS} to weight the losses. Differently to \cite{Siarohin_2019_NeurIPS}, we turned adversarial training on from the beginning of training for Frontalizer models. Indeed, while for the FOM baseline adversarial fine-tuning was experimentally better than direct training in the dense setting, for our other models we performed adversarial training from scratch.

\paragraph{Additional losses for Sup+Unsup models.}

We tried to add Jacobians as in \cite{Siarohin_2019_NeurIPS} but it did not change performance, cf. Table \ref{tab:jac_ablation}.

\subsection{Additional results}

\subsubsection{Cross-reenactment} 
\begin{figure}[h]
  \centering
  \resizebox{1.0\linewidth}{!}{
    \setlength{\tabcolsep}{1pt}
  \begin{tabular}{ccc}
Source S1 & Source S2 & Driving \\
  \includegraphics[height=0.32\linewidth]{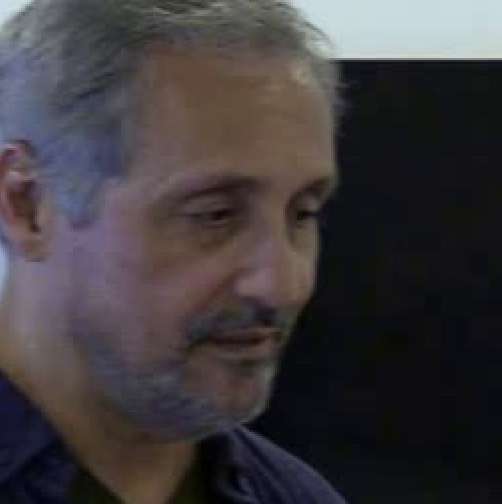}& \includegraphics[height=0.32\linewidth]{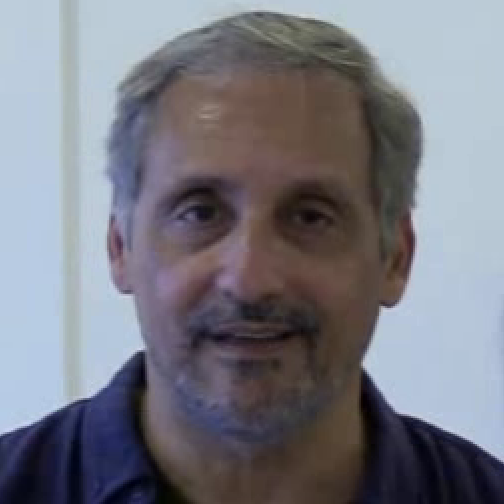}& 
  \includegraphics[height=0.32\linewidth]{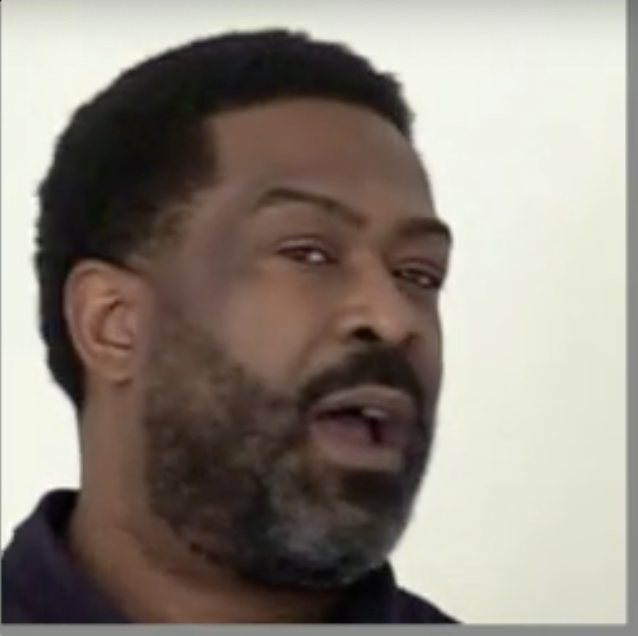}\\
   FOM with S1 & FOM with S2 & Ours \\
  \includegraphics[height=0.32\linewidth]{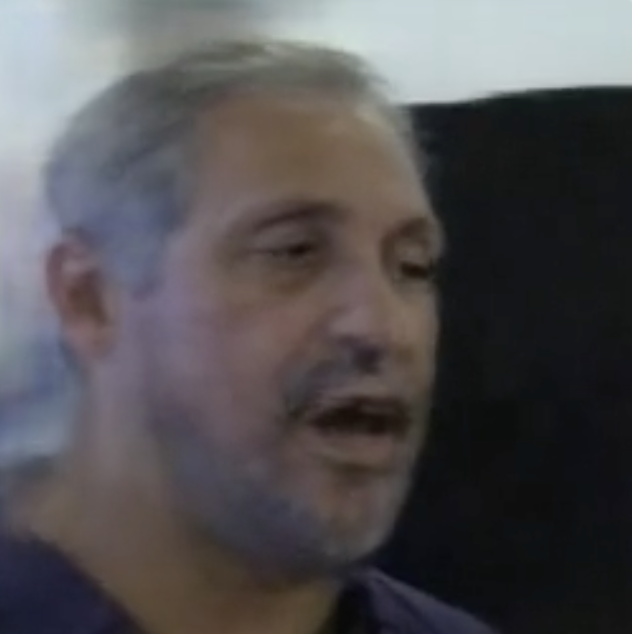}&
    \includegraphics[height=0.32\linewidth]{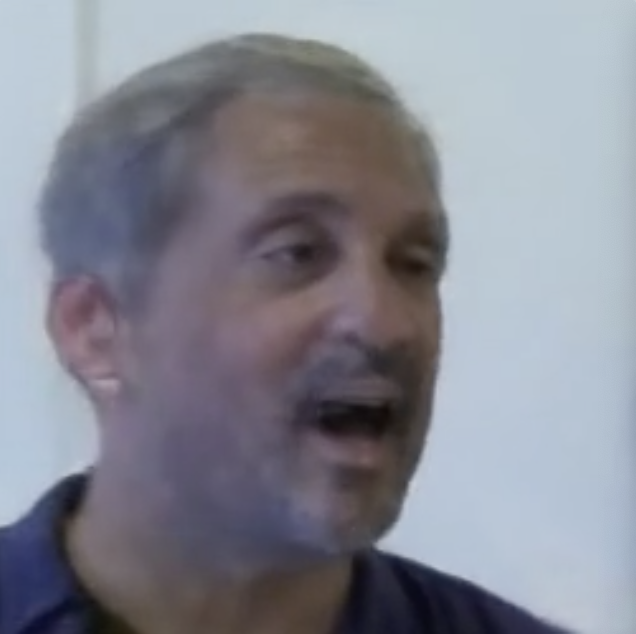}&
      \includegraphics[height=0.32\linewidth]{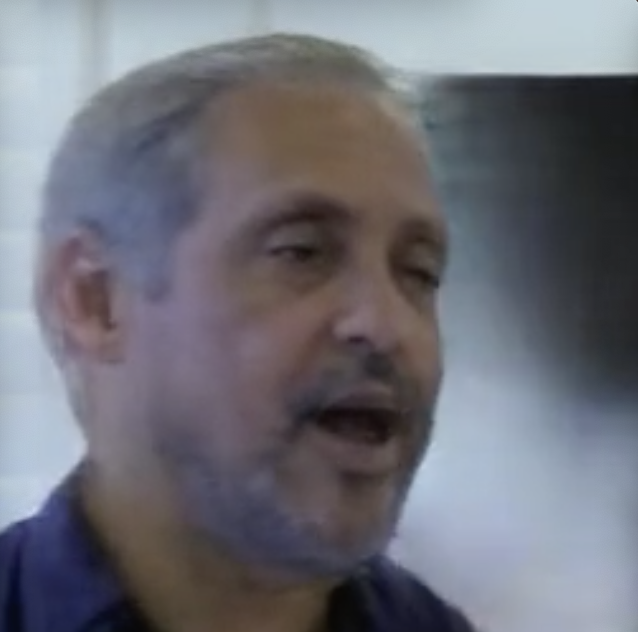}
  \end{tabular}}
     \caption{Cross reenactement: Face animation given a few source frames using landmarks from a driving video.}
\label{fig:cross}
\end{figure}

We finally show that our approach is able to provide cross-reenactment results as illustrated in Figure~\ref{fig:cross}.

\begin{table*}[ht]
    \centering
     \resizebox{1\textwidth}{!}{ 
    \begin{tabular}{lccccccccc}
    \toprule
    & \multicolumn{4}{c}{\textbf{DFDC-rotations}} & \multicolumn{4}{c}{\textbf{VoxCeleb}}\\
      & \small LPIPS $\downarrow$  & \small NME $\downarrow$ & \small CSIM $\uparrow$ & Expr $\uparrow$ & \small LPIPS $\downarrow$  & \small NME $\downarrow$ & \small CSIM $\uparrow$ & Expr $\uparrow$\\
    \midrule
          Frontalizer sup+unsup + free expr {\bf with Jacobians} & {0.113}	{\tiny $\pm$0.03}&	{0.467}{\tiny $\pm$	0.38}&	0.789{\tiny $\pm$0.06}&{ 0.897}{\tiny $\pm$	0.05} & 0.140{\tiny $\pm$0.04}&0.256{\tiny {$\pm$0.04}}&0.781{\tiny $\pm$0.06}&	0.910{\tiny $\pm$0.05}\\ %20211025/facegen.UnFlZUiFiq bs 30
        Frontalizer sup+unsup + free expr, {\bf no Jacobians} & {0.115}	{\tiny $\pm$0.03}&	{ 0.463}{\tiny $\pm$	0.42}&	0.787{\tiny $\pm$0.06}&{ 0.894}{\tiny $\pm$	0.05} & 0.147{\tiny $\pm$0.04}&0.247{\tiny {$\pm$0.04}}&0.779{\tiny $\pm$0.06}&	0.909{\tiny $\pm$0.04}\\
        \bottomrule 
    \end{tabular}}
    \caption{Ablation study on the impact of Jacobians in the dense Sup Unsup Expr model.}
    \label{tab:jac_ablation}
\end{table*}

\subsubsection{Head shape and expression preservation}

\begin{figure*}[htb] \RawFloats 
\centering
\begin{minipage}{.5\textwidth}
  \centering
     \includegraphics[width=0.3\textwidth,height=0.3\textwidth]{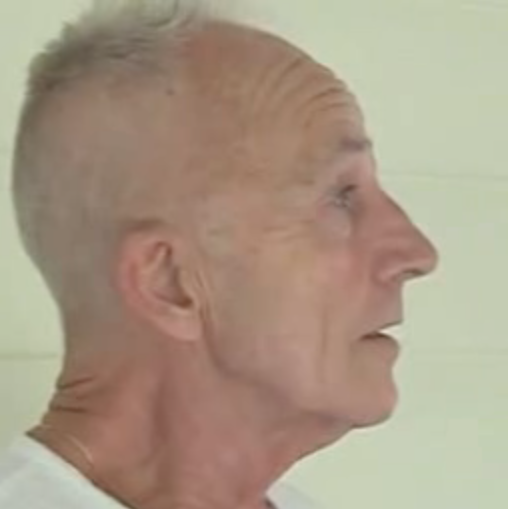} 
      \includegraphics[width=0.3\textwidth,height=0.3\textwidth]{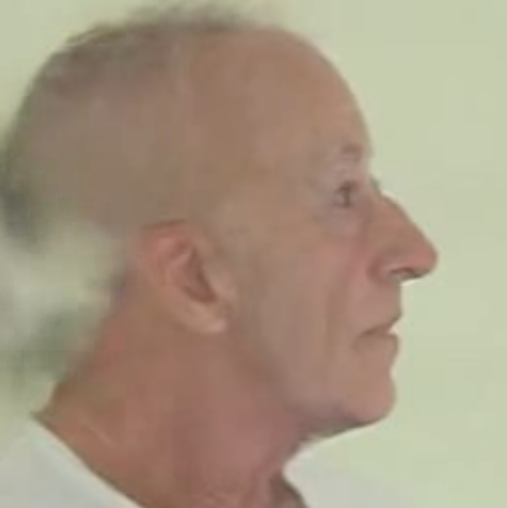} 
       \includegraphics[width=0.3\textwidth,height=0.3\textwidth]{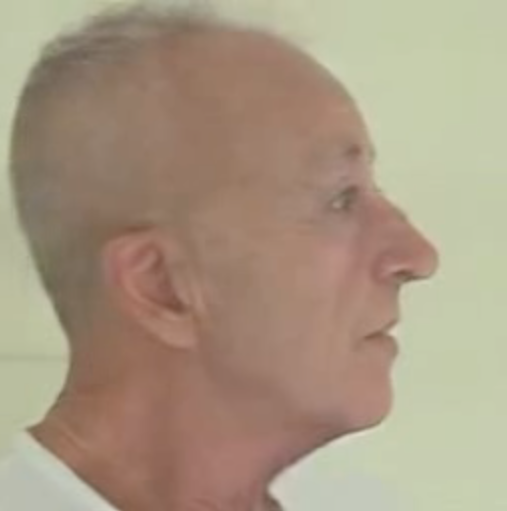} \\
        \includegraphics[width=0.3\textwidth,height=0.3\textwidth]{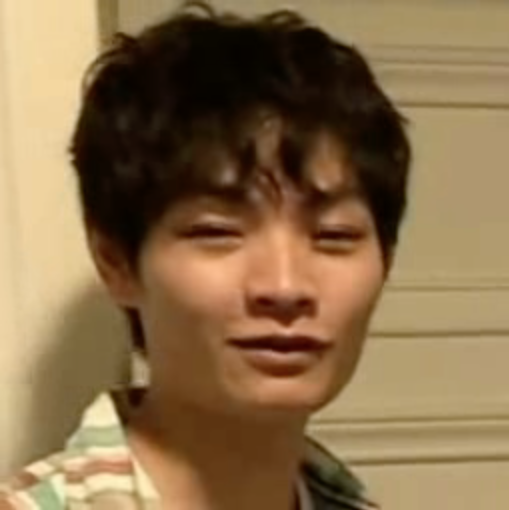} 
      \includegraphics[width=0.3\textwidth,height=0.3\textwidth]{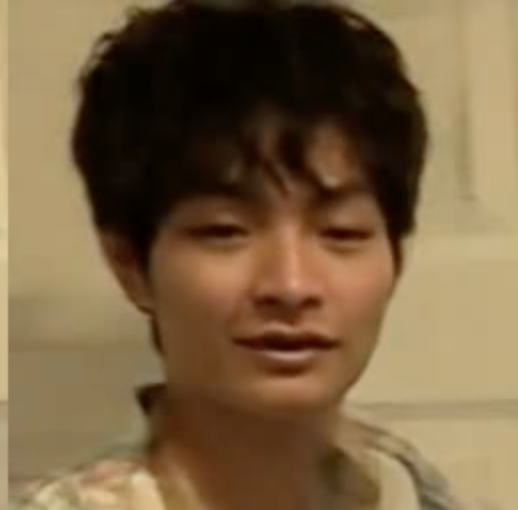} 
       \includegraphics[width=0.3\textwidth,height=0.3\textwidth]{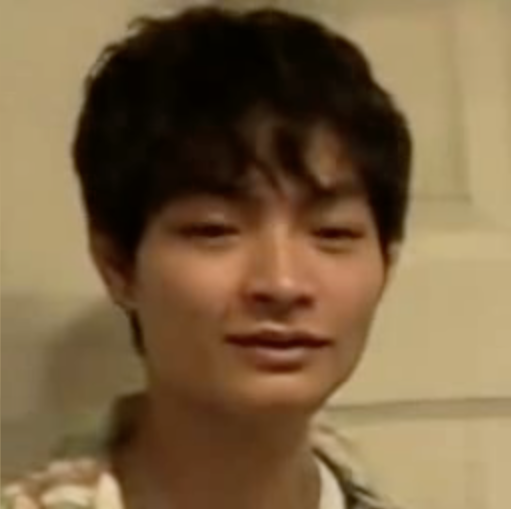}
    \captionof{figure}{Head shape improvements on top of Frontalizer sup.\\ 
    Left: original; Middle Frontalizer sup. Right: Frontalizer sup+unsup.}
    \label{fig:head}
\end{minipage}%
~~
\begin{minipage}{.48\textwidth}
  \centering
   \includegraphics[width=0.95\textwidth]{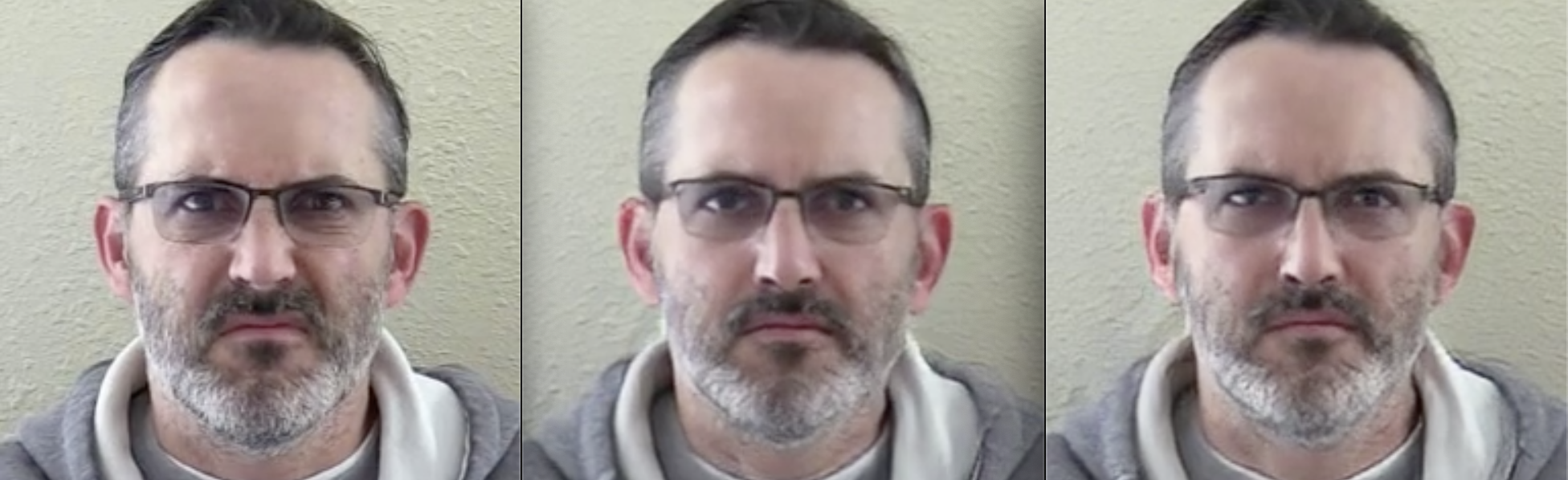} \\
     \includegraphics[width=0.95\textwidth]{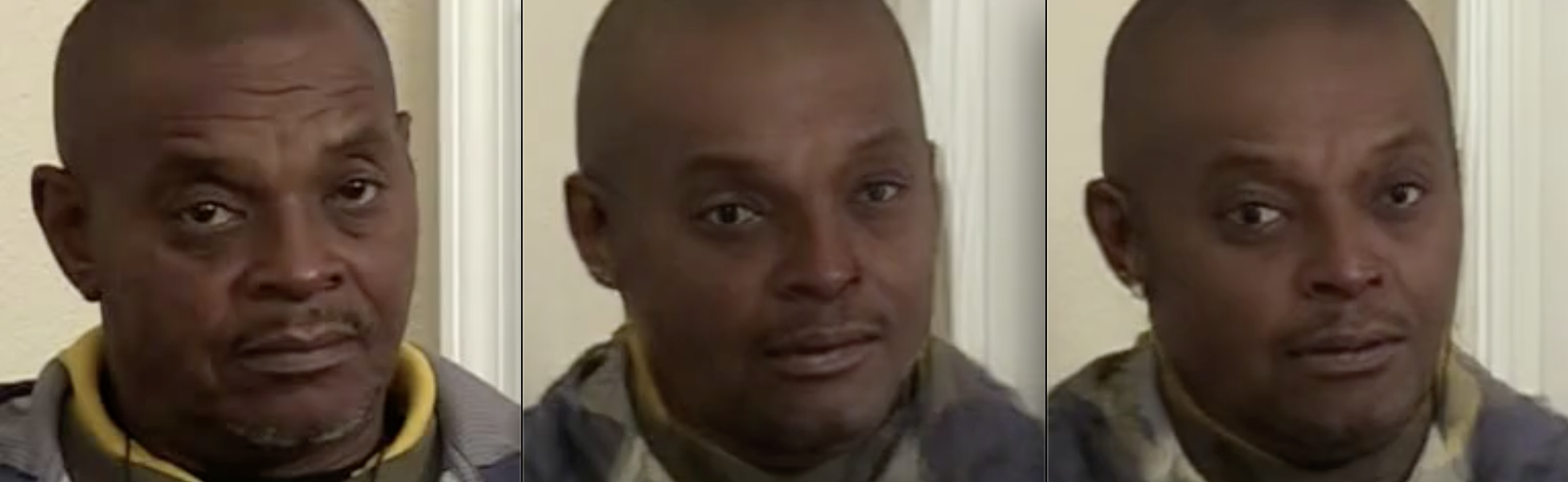} \\  
    \captionof{figure}{Expression improvements on top of Frontalizer unsup+sup. Left: original; Middle Frontalizer baseline; right: baseline with free expression conditioning.}
    \label{fig:free_expr_app}
\end{minipage}
\end{figure*}

Figure \ref{fig:head} provides examples of improvements of the Frontalizer \Supunsup~model over the Frontalizer \Sup~ dense model. 

Figure \ref{fig:free_expr_app} shows examples of additional improvements brought by expression conditioning on top of the Frontalizer \Supunsup~dense model.

\subsubsection{Ablation study on mobile models}

\begin{figure*}[htb]
\centering
~~~~~~ Driving ~~~~~~~~~~~~~~~~  ~~~~~~~~~~FOM~~~~~~~~ ~~~~~~~~~~ ~~~~~~~~Front. S~~~~~~~~~~~ ~~~~~~~~~~~~~~~~~~~~Front. S+U ~~~~~~~~~~~~~~~~~~~~~~Front S+U+E~~~~~\\
     \includegraphics[width=1\textwidth]{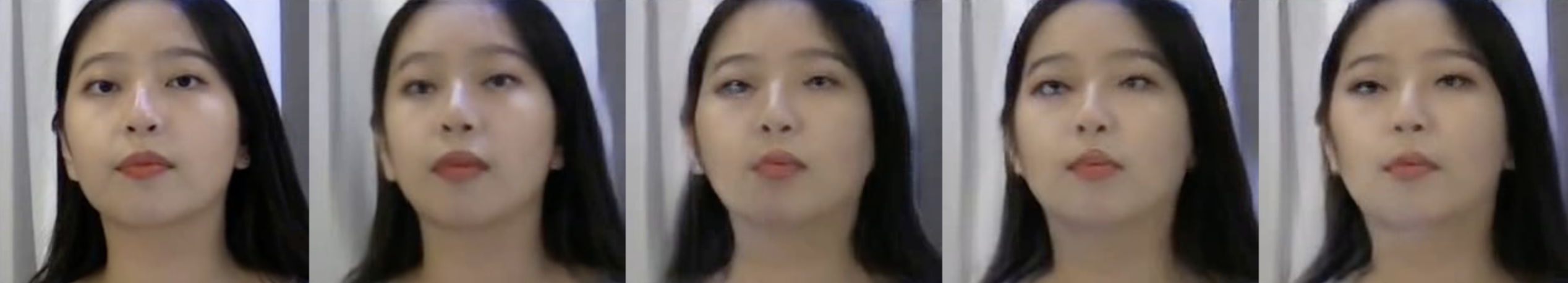} \\
     \includegraphics[width=1\textwidth]{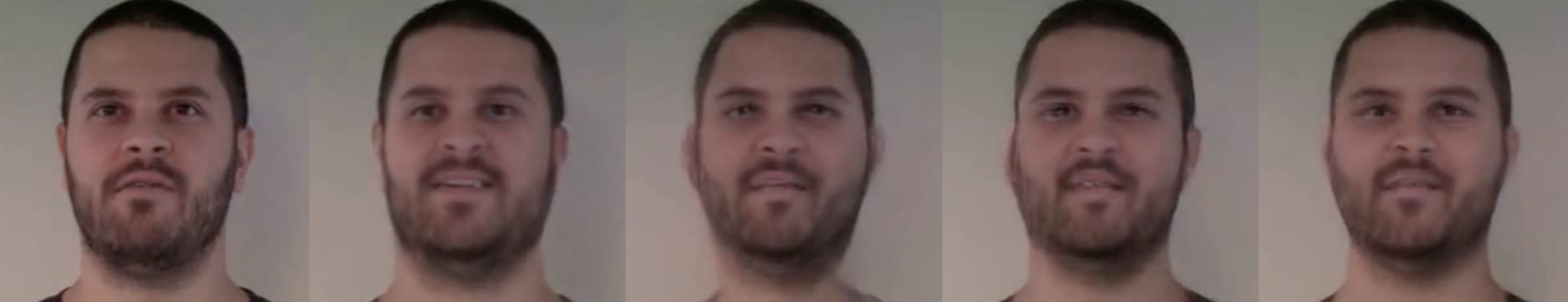} \\
     \includegraphics[width=1\textwidth]{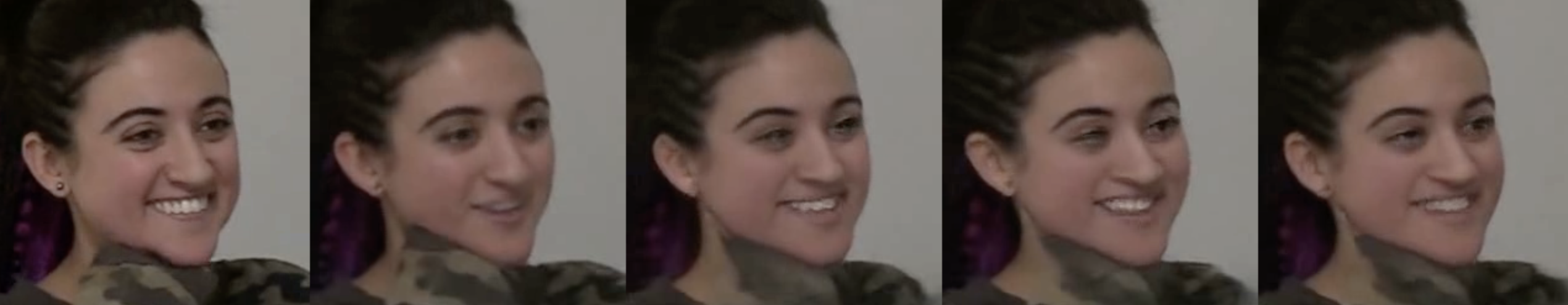} 
    \caption{Mobile ablation results. Expression conditioning improves eye rendering.}
    \label{fig:mobile2}
\end{figure*}

Metrics using expression in mobile models are provided in Table~\ref{tab:ExpressionMobile}. From this table, it seems that expression conditioning does not bring much to the Sup+ Unsup Frontalizer. Qualitatively, we observe improvement of eye rendering, see Figure~\ref{fig:mobile2}.

\begin{table*}[htb]
    \centering
     \resizebox{1\textwidth}{!}{ 
    \begin{tabular}{lccccccccc}
    \toprule
    & \multicolumn{4}{c}{\textbf{DFDC-rotations}} & \multicolumn{4}{c}{\textbf{VoxCeleb}}\\
      & \small LPIPS $\downarrow$  & \small NME $\downarrow$ & \small CSIM $\uparrow$ & Expr $\uparrow$ & \small LPIPS $\downarrow$  & \small NME $\downarrow$ & \small CSIM $\uparrow$ & Expr $\uparrow$\\
    \midrule
          Ours (sup+unsup) & {\bf 0.148} {\tiny $\pm$0.003}&  0.585 {\tiny $\pm$0.044}& {\bf 0.723} {\tiny $\pm$0.006}& {\bf 0.865} {\tiny $\pm$0.005}& {\bf 0.165} {\tiny $\pm$0.002}& {\bf 0.275} {\tiny $\pm$0.003}& {\bf 0.747} {\tiny $\pm$0.003}& {\bf 0.893} {\tiny $\pm$0.002}\\
           Ours (sup+unsup+expr) &  0.151 {\tiny $\pm$0.003}& {\bf 0.565} {\tiny $\pm$0.044}& { 0.718} {\tiny $\pm$0.006}& {\bf 0.865} {\tiny $\pm$0.005}& \bf 0.167 {\tiny $\pm$0.002}& {\bf 0.273} {\tiny $\pm$0.003}& {\bf 0.748} {\tiny $\pm$0.003}& { 0.886} {\tiny $\pm$0.002}\\
        \bottomrule 
    \end{tabular}}
    \vspace{-2ex}
    \caption{Ablation study on the impact of expressions in the mobile Sup Unsup  model (no Jacobians).
    }
    \label{tab:ExpressionMobile}
\end{table*}

\end{document}